\documentclass[lettersize,journal]{IEEEtran}
\usepackage{amsmath,amsfonts}
\usepackage[table]{xcolor}  
\usepackage{multirow}
\usepackage{algorithmic}
\usepackage{algorithm}
\usepackage{array}
\usepackage{booktabs}
\usepackage[caption=false,font=normalsize,labelfont=sf,textfont=sf]{subfig}
\usepackage{textcomp}
\usepackage{diagbox}
\usepackage{stfloats}
\usepackage{url}
\usepackage{verbatim}
\usepackage{graphicx}
\usepackage{cite}
\usepackage{pifont}
\usepackage{hyperref}
\usepackage{xcolor}
\usepackage{makecell}
\begin{document}

\title{TUNI: Unifying Pre-training and Fine-tuning with Modality-Aware Mutual Learning and Rectification for RGB-T Semantic Segmentation}

\author{Xiaodong Guo, Xianda Guo, Tong Liu, Zhihong Deng, Yanlun Peng, Xiang Li, 

Wujie Zhou,~\IEEEmembership{Senior Member,~IEEE}
\thanks{This work was supported by National Natural Science Foundation of China under Grant 62476026. 
(Corresponding authors: Tong Liu, Wujie Zhou, Xianda Guo).

Xiaodong Guo, Zhihong Deng, Tong Liu and Xiang Li are with the School of Automation, Beijing Institute of Technology, Beijing 100081, China (E-mail: guoxd@bit.edu.cn, liutong2002@bit.edu.cn).

Xianda Guo is with the School of Computer Science, Wuhan University, Wuhan, China.

Yanlun Peng is with the Great Wall Motor, Shanghai, China.

Wujie Zhou is with the School of Information and Electronic Engineering, Zhejiang University of Science and Technology, Hangzhou 310023, China.

© 2026 IEEE. Personal use of this material is permitted. Permission from IEEE must be obtained for all other uses, in any current or future media, including reprinting/republishing this material for advertising or promotional purposes, creating new collective works, for resale or redistribution to servers or lists, or reuse of any copyrighted component of this work in other works.

}}



\maketitle

\begin{abstract}
RGB-thermal (RGB-T) semantic segmentation improves the environmental perception of autonomous platforms in challenging conditions. Prevailing RGB-T segmentation frameworks suffer from suboptimal multi-modal feature extraction and fusion, unbalanced modality dependency, and inadequate utilization of thermal information. To address these challenges, we propose TUNI, a unified pre-training and fine-tuning framework for efficient and real-time RGB-T semantic segmentation. It pre-trains an RGB-T encoder that incorporates an RGB-T local module that selectively emphasizes salient consistent and distinct local features across modalities, thereby integrating cross-modal feature extraction and fusion in a unified manner. To alleviate the modality bias issue during RGB-T pre-training, modality-inverted contrastive mutual learning is introduced to enable knowledge exchange between two RGB-dominated and thermal-dominated encoders. In the fine-tuning phase, modality rectification learning fully exploits residual thermal information by focusing on correct yet divergent prediction regions between two modality-specific decoders. We further develop three TUNI variants, covering lightweight, balanced, and high-performance requirements. Extensive experiments on five RGB-T semantic segmentation datasets demonstrate that TUNI achieves superior accuracy, generalization, and compactness compared with 15 state-of-the-art models. The code is available at https://github.com/xiaodonguo/TUNI-v2. 
\end{abstract}


\section{Introduction}
\IEEEPARstart{T}{he} intelligence of autonomous platforms (e.g., mobile robots, drones, autonomous cars) relies heavily on the perception and understanding of their surroundings. This can be achieved by RGB-thermal (RGB-T) semantic segmentation, which performs pixel-level classification on images utilizing RGB and thermal information captured by visible-light and long-wave thermal cameras, respectively. The RGB information provides rich color and texture cues, while the thermal information captures heat distribution, serving as a complement in challenging scenarios, such as low-illumination \cite{MFNet, tcsvt1}, glare \cite{FMB, tcsvt2} and low color contrast \cite{SUS, CART}.

\begin{figure*}[t]
\centering
\includegraphics[width=\textwidth]{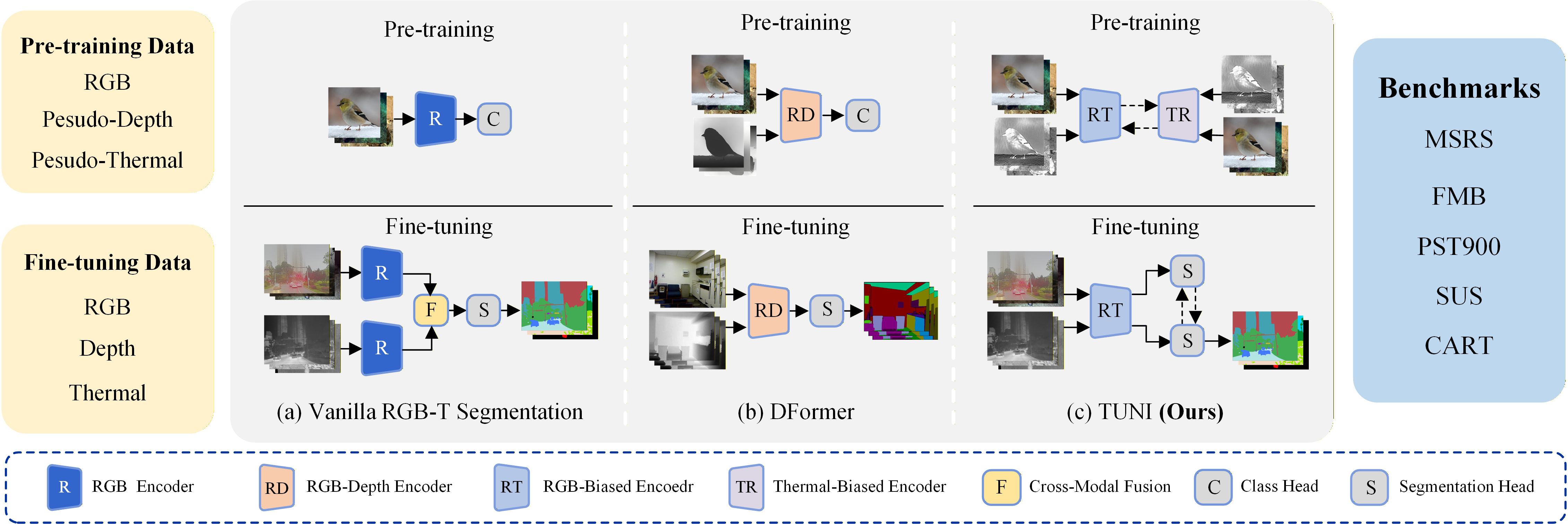}
\caption{Three RGB-T/RGB-D semantic segmentation frameworks: (a) Vanilla RGB-T segmentation framework. It pre-trains an RGB encoder on ImageNet-1K, and fine-tunes it with the cross-modal feature fusion module and segmentation head on segmentation datasets. (b) DFormer. It pre-trains an RGB-D encoder with RGB and pseudo-depth images from ImageNet-1K, therefore performs multi-modal feature extraction and cross-modal feature fusion simultaneously. (c) TUNI. It first pre-trains a dedicated RGB-T encoder with RGB and pseudo-thermal images from ImageNet-1K, incorporating modality-aware mutual learning to alleviate modality bias. The modality rectification learning is then introduced during the fine-tuning phase to make full use of the thermal information.}
\label{fig1}
\end{figure*}

Prevailing RGB-T semantic segmentation models (Fig. \ref{fig1}(a)) typically follow a framework that pre-trains RGB encoders \cite{segformer, ResNet} on ImageNet-1K \cite{ImageNet} and then fine-tunes them by integrating additional cross-modal feature fusion modules and segmentation head on segmentation datasets \cite{MFNet, CART, PST900, SUS}. To improve segmentation accuracy, research focuses on improving cross-modal feature fusion, primarily consisting of methods based on Convolution-Attention \cite{GMNet, CLNet, tcsvt3}, Transformer-Attention\cite{CMX, CMNext, SUS, MMSMCNet}, and Mamba \cite{sigma, CM-SSM, MDNet}. {Although effective, this framework suffers from two primary limitations, including 1) considerable redundancy introduced by duplicated feature extraction structures, and 2) insufficient utilization of thermal information due to the absence of thermal cues during pre-training.}

{To alleviate structural redundancy, recent works attempt to simplify the thermal branch by replacing or lightweighting the thermal encoder, such as using multi-modal prompt generation modules or asymmetric lightweight encoders \cite{Dong, kang, Sun}. However, these approaches either restrict cross-modal interactions to limited layers or rely on encoders pre-trained on RGB data, which limits their ability to effectively capture thermal-specific characteristics. More importantly, the second limitation remains largely unresolved. Due to the lack of large-scale thermal datasets, it is inherently difficult to pre-train a dedicated thermal encoder, which further constrains the effective utilization of thermal information. Therefore, a critical question arises: how to design a unified pre-training and fine-tuning framework that can mitigate structural redundancy and effectively exploit thermal information across both stages.}

In RGB-Depth (RGB-D) semantic segmentation tasks \cite{tcsvt3, tcsvt4}, Yin et al. \cite{DFormer} propose DFormer, which approaches the aforementioned issues from a novel perspective, as shown in Fig. \ref{fig1}(b). The DFormer utilizes a depth estimation network to generate RGB-D image pairs from ImageNet-1K, and directly pre-trains an RGB-D encoder that simultaneously performs multi-modal feature extraction and cross-modal feature fusion. In the RGB-D encoder, the parameters of the depth branch are half that of the RGB to reduce redundancy. {However, several challenges remain when adopting DFormer to RGB-T semantic segmentation tasks. First, the heterogeneity between thermal and RGB images is greater than that between depth and RGB images, increasing the difficulty of cross-modal feature fusion. Second, reducing the parameters of thermal branch by half may cause over-reliance on the RGB modality during the pre-training process, which we refer to as modality bias. Third, the DFormer utilizes only the RGB branch for segmentation during fine-tuning, as it assumes that the depth information has already been sufficiently integrated within the encoder. However, due to residual connections, the thermal branch still retains substantial information that deserves to be utilized.}

To address the aforementioned challenges, we propose a novel framework, named TUNI, for efficient and real-time RGB-T semantic segmentation. {The key idea of TUNI is to enable effective large-scale RGB-T pre-training with consistent thermal utilization across both pre-training and fine-tuning stages, while mitigating structural redundancy.} As shown in Fig. \ref{fig1}(c), the training of TUNI includes two steps: RGB-T pre-training and fine-tuning. In the pre-training phase, an RGB-T image translation model \cite{translation} is utilized to generate RGB and pseudo-thermal image pairs from ImageNet-1K. {Considering the heterogeneity between RGB and thermal information,} we introduce an RGB-T local module (R-T LM) that employs adaptive cosine similarity to selectively emphasize salient consistent and distinct local features across RGB-T modalities. {To further mitigate modality bias,} we propose a \textbf{M}odality-\textbf{I}nverted \textbf{C}ontrastive \textbf{M}utual \textbf{L}earning (MI-CML) strategy by constructing an additional T-RGB encoder, in which the parameters of the thermal branch are twice those of the RGB branch, to learn thermal-dominated features. By integrating contrastive learning and mutual learning, effective knowledge exchange between the two encoders is achieved in the representation space. {Finally, to fully exploit residual thermal information during fine-tuning,} we design a \textbf{M}odality \textbf{R}ectification \textbf{L}earning (MRL) decoder, which focuses on learning from correct yet divergent regions between the predictions of two modality-specific decoders to extract valuable residual thermal cues. 

{Unlike conventional two-stage pipelines where pre-training and fine-tuning are designed independently, our framework enforces a unified learning paradigm across both stages. Compared with DFormer, the proposed R-T LM, MI-CML, and MRL operate at the architectural level, pre-training stage, and fine-tuning stage, respectively, forming a progressive design that incrementally enhances the framework’s ability to effectively exploit thermal information.} Our main contributions can be summarized as follows:
\begin{itemize}
\item[$\bullet$] We propose TUNI, a unified pre-training and fine-tuning framework for efficient and real-time RGB-T semantic segmentation. With RGB and pseudo-thermal images from ImageNet-1K, it pre-trains an RGB-T encoder consisting of an R-T LM that employs adaptive cosine similarity to selectively emphasize salient consistent and distinct local features across RGB-T modalities.
\item[$\bullet$] We propose the MI-CML to alleviate the modality bias caused by unbalanced parameter settings between the RGB and thermal branches within the RGB-T encoder during the pre-training phase. The MI-CML constructs an additional thermal-dominated encoder, and facilitates knowledge exchange with the RGB-dominated encoder in the representation space.
\item[$\bullet$] We propose an MRL decoder to fully utilize the thermal information during the fine-tuning phase. It focuses on learning from the correct yet divergent regions in the predictions between two modality-specific decoders to extract valuable residual thermal information.
\item[$\bullet$] We develop three encoder versions, namely TUNI-T, TUNI-S and TUNI-B, which correspond to lightweight, balanced and high-performance models, respectively. Experiments conducted on five RGB-T semantic segmentation datasets demonstrate that our models achieve superior generalization and compactness compared with 15 state-of-the-art (SOTA) models.
\end{itemize}

In this extended version for the submission of TCSVT, we have made significant improvements to our work on the conference version, which has been originally accepted by ICRA 2026 \cite{TUNI}. The enhancements are as follows: (1) we propose MI-CML, a novel multi-modal pre-training strategy. It facilitates knowledge exchange between two RGB-dominated and thermal-dominated encoders in the representation space via contrastive learning and mutual learning, thereby alleviating the modality bias during the pre-training phase. (2) we propose an MRL decoder to fully exploit thermal information during the fine-tuning phase. This residual information is shown to effectively enhance segmentation accuracy, yet is ignored in the DFormer framework and our conference version. (3) we further explore the trade-off between accuracy and model lightweightness by introducing two encoder variants, TUNI-T and TUNI-B, which respectively target lightweight deployment and high-performance requirements. (4) We have expanded our experiments by including 6 SOTA models: SGFNet \cite{SGFNet}, MCNet-T \cite{SUS}, Sigma \cite{sigma}, CM-SSM \cite{CM-SSM}, FGDNet-S \cite{FGDNet-S}, and TUNI \cite{TUNI}. In addition, we conduct additional experiments on two public datasets, MSRS \cite{MSRS} and SUS \cite{SUS}, to further verify the generalization capability of our model.

\section{Related Work}
\subsection{RGB-T Semantic Segmentation}
Traditional semantic segmentation \cite{FCN,PSANet,PSPNet,Deeplabv3} performs pixel-level prediction on RGB images, aiding autonomous platforms in understanding environmental semantics. {To enhance global contextual reasoning, CDGR \cite{CDGR} further incorporates language as an auxiliary modality by introducing a cross-modal dual graph reasoning framework that performs joint reasoning over spatial dependencies and semantic relationships via visual–language graphs.} However, RGB images suffer from the loss or confusion of object features under adverse conditions, such as low-illumination and glare. RGB-T semantic segmentation incorporates thermal information that is insensitive to light, demonstrating robustness under such adverse conditions and becoming an alternative.

The prevailing RGB-T semantic segmentation frameworks typically employ the pre-trained encoder (e.g., ResNet \cite{ResNet}, Mit\cite{segformer}) to extract RGB and thermal features, and focus on improving the efficiency of cross-modal feature fusion. Early cross-modal feature fusion strategies are based on Convolution-Attention. According to the characteristics of features at different dimensions, Zhou et al. \cite{GMNet} utilize spatial attention and channel attention for low-dimensional and high-dimensional features, respectively. Zhou et al. \cite{MFFENet} develop a compact version of ASPP \cite{Deeplabv3} to modify it to RGB-T features. Chen et al. \cite{SAGate} propose an SA-Gate module to realize RGB-D feature separation and feature aggregation, proven to be effective in RGB-T feature fusion \cite{CMX}. Guo et al. \cite{CLNet} propose an MHAI module to gradually combine complementary and common RGB-T features in distinct receptive fields. Wang et al. \cite{SGFNet} enhance the fused features with cross-level information and semantic information via row pooling and 1D position encoding. In addition to Convolution-Attention strategies, Transformer-Attention fusion has also been widely adopted, which models long-range dependencies between RGB-T features. Pang et al. \cite{CAVER} propose the intra-modal self-attention and inter-modal cross-attention to construct long-range dependencies inside and between modalities. Zhou et al. \cite{MMSMCNet} propose a modal memory sharing module that memorizes the attention along the vertical and horizontal directions. Guo et al. \cite{SUS} construct cross-attention across both channel and spatial dimensions. Zhang et al. \cite{CMX, CMNext} divide the fusion phase with cross-modal features rectification and cross-modal feature fusion stages, which combines the Convolution-Attention and Transformer-Attention strategies. Although Transformer-based strategies can capture long-range dependencies, the significant computational overhead cannot be ignored, which scales quadratically with image size. To reduce the computational burden, state space model has been introduced to cross-modal feature fusion, which we refer to as Mamba-Attention strategies. Peng et al. \cite{Fusionmamba} propose a FusionMamba block, serving as a plug-and-play information fusion module. Wan et al. \cite{sigma} utilize the VMamba \cite{VMamba} as the encoder and employ a cross-multiplication mechanism to enhance the features with one another. Guo et al. \cite{CM-SSM} propose a CM-SS2D module to construct cross-modal vision sequence and obtain the hidden state of one modality through the other. However, the above framework separates the encoder and cross-modal feature fusion, which has been demonstrated to be suboptimal \cite{DFormer}.

In RGB-D semantic segmentation, Yin et al. \cite{DFormer} propose the DFormer, which pre-trains an RGB-D encoder that performs multi-modal feature extraction and cross-modal feature fusion simultaneously, thereby improving the accuracy and compactness of the model. {However, directly adopting DFormer for RGB-T segmentation is suboptimal due to the heterogeneity of the thermal modality, the inherent inter-modal parameter imbalance, and the neglect of residual thermal information. To this end, we propose the TUNI framework, consisting of an R-T LM that performs delicate local RGB-T feature fusion and MI-CML that alleviates the modality bias in the pre-training phase. Furthermore, the MRL decoder is introduced to utilize the thermal residual information in the fine-tuning stage, which is ignored by DFormer.}

\subsection{Mutual Learning}
Unlike unidirectional knowledge transfer of knowledge distillation\cite{KD} that transfers knowledge from teacher models to student models, mutual learning\cite{dml} facilitates the exchange of knowledge between student models. Zhang et al. \cite{dml} initially demonstrated that students initialized with different random parameters exhibit distributional divergencies, which facilitates mutual learning among diverse students. Guo et al.\cite{onlinekd} ensemble the soft targets produced by all students as learning targets to gain extra information. Zhang et al.\cite{ckd} involved students in the knowledge distillation process to assist the teacher in transferring knowledge. Except for the  knowledge transfer of logit prediction, Gou et al.\cite{ckdmt} introduce the relation-based knowledge transfer to enrich the transferred knowledge. Yang et al.\cite{MCL} integrate the mutual learning with contrastive learning \cite{SCL}, transferring the contrastive distributions among a cohort of networks. Zhou et al. \cite{RCNet} introduce the geometric resonance adversarial learning to exchange the orientation-sensitive feature extracted by Mamba-based mechanism and the macro structural feature aggregated in the frequency domain. Wu et al.\cite{RCL} propose rectified logit-wise collaborative learning and class-aware feature-wise collaborative learning to facilitate knowledge exchange between CNN-based and Transformer-based models at the logit and feature levels, respectively. 

{Despite progress in applying mutual learning across different model architectures, its exploration across diverse modalities and in the context of pre-training and fine-tuning transfer remains unexplored. In this paper, we propose MI-CML, which generates modality distribution divergencies through modality inversion and utilizes contrastive mutual learning to mitigate modality bias during the pre-training phase. In addition, we introduce MRL to restrict mutual learning to regions where both modalities make correct yet different predictions, thereby providing more informative
supervision for cross-modal knowledge transfer.}

\begin{figure*}[t]
\centering
\includegraphics[width=\textwidth]{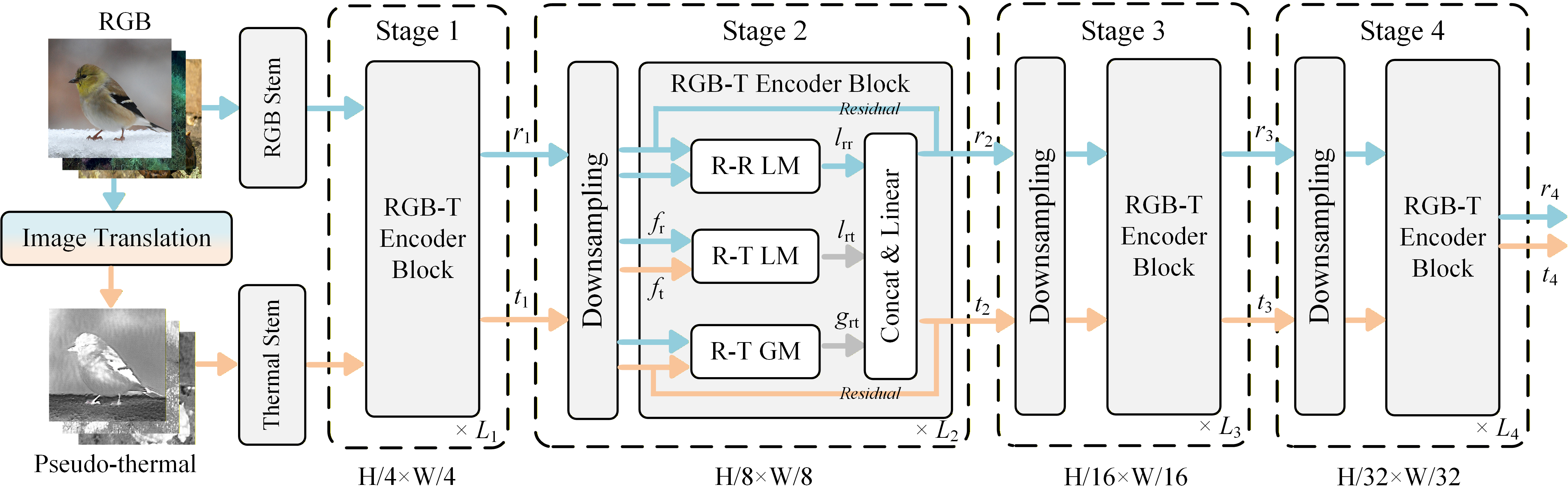}
\caption{Illustration of the TUNI encoder. The TUNI encoder consists of multiple stacked RGB-T encoder blocks, each of which includes R-R LM, R-T LM, and R-T GM. It performs multi-modal feature extraction and cross-modal feature fusion simultaneously, achieving modality-specific feature extraction while improving fusion efficiency.}
\label{fig2}
\end{figure*}

\section{Proposed Method}
\subsection{Overview}
The TUNI framework comprises three key components: TUNI encoder, pre-training phase, and fine-tuning phase. The TUNI encoder is specifically designed to extract RGB-T features and perform cross-modal feature fusion simultaneously. Within the encoder, an R-T LM is introduced to enable delicate local feature fusion between RGB and thermal modalities, taking into account the unique characteristics of each modality. During the pre-training phase, the TUNI encoder is trained on ImageNet-1K\cite{ImageNet} with the help of an RGB-T image translation network\cite{translation}, which generates paired RGB and pseudo-thermal images. In this phase, MI-CML is incorporated to mitigate the modality bias arising from modality imbalances among the RGB and thermal branches within the encoder. In the fine-tuning phase, the pre-trained TUNI encoder is adapted to RGB-T semantic segmentation datasets with an MRL decoder, fully exploiting the residual information from the thermal branch.
\subsection{TUNI Encoder}
The overall architecture of TUNI encoder is illustrated in Fig. \ref{fig2}. It consists of multiple blocks, each comprising three sub-modules: RGB-RGB local module (R-R LM), RGB-T global module (R-T GM) and R-T LM. These blocks are organized into four stages $\left\{L_{1}, L_{2}, L_{3}, L_{4}\right\}$, where $L_{{n}}$ indicates the number of blocks. Through downsampling, the feature resolutions at each stage are H/4 × W/4, H/8 × W/8, H/16 × W/16, and H/32 × W/32, where H and W denote the height and width of the original input image. Since RGB images carry richer information than thermal images, the channel dimension of thermal features is set to half that of the RGB features. Consequently, the thermal branch of the TUNI encoder has only half the parameters of the RGB branch, reducing the overall model complexity.

\begin{figure}[t!]
      \centering
      \includegraphics[width=0.9\columnwidth]{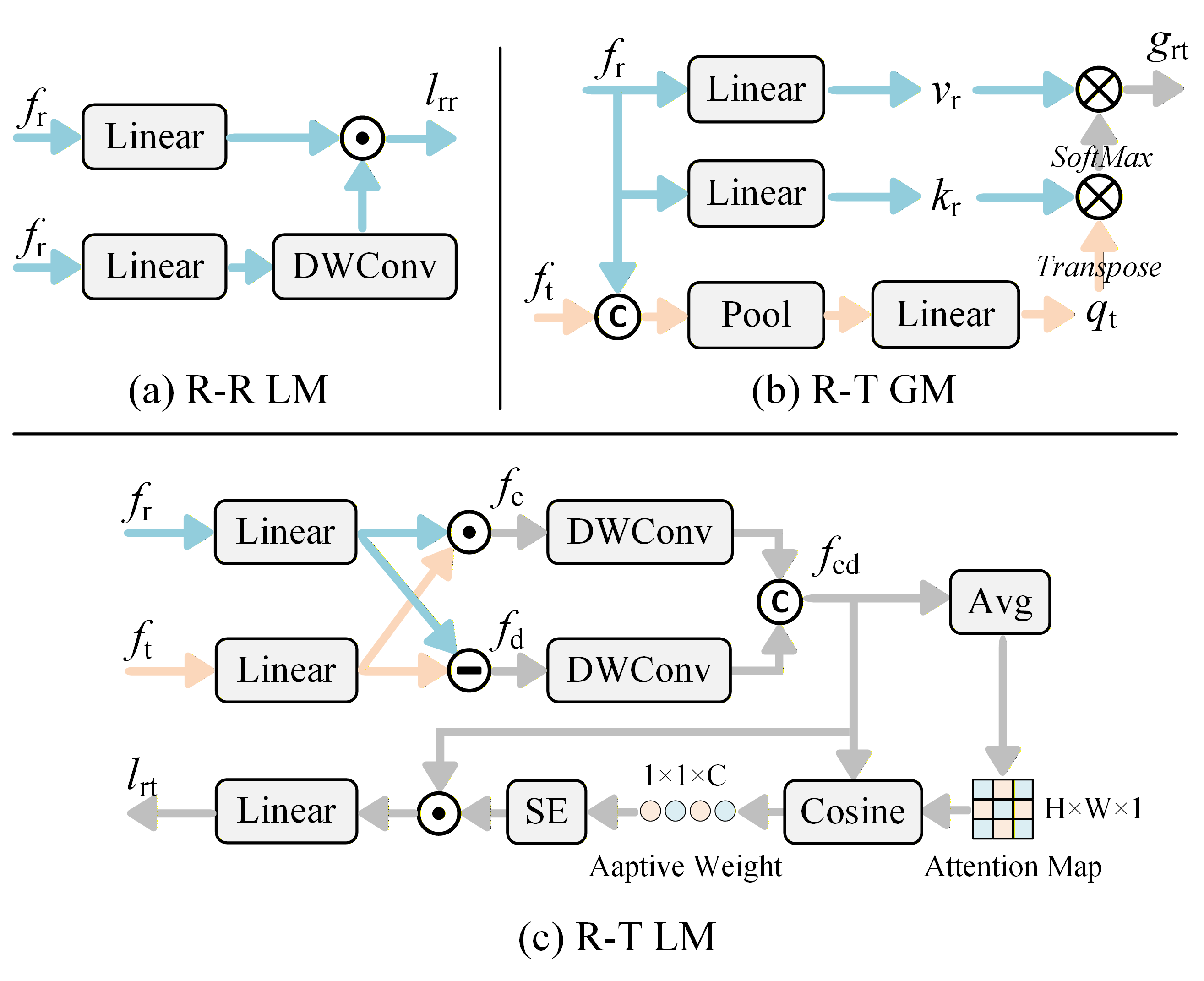}
      \caption{Three sub-modules in the TUNI encoder: (a) R-R LM, (b) R-T GM, (c) R-T LM.}
      \label{fig3}
\end{figure}

\textit{\textbf{R-R LM and R-T GM:}} Inherited from DFormer, the R-R LM and R-T GM implement local RGB feature extraction and global RGB-T feature fusion, respectively. The internal structure of the R-R LM is shown in Fig. \ref{fig3}(a) (the normalization layer is omitted for convenience). In addition to basic linear layers, a depthwise separable convolution layer is employed to further extract local features. This module can be formulated as:

\begin{equation}
l_{\text{rr}} = Linear(f_{{r}}) \odot DWConv(Linear(f_{{r}}))
\end{equation}
where $f_{r}$ and $l_\text{rr}$ denote RGB and local-enhanced RGB features, respectively. $Linear(.)$ and $DWConv(.)$ denote the linear and depthwise separable convolution layers, respectively. $\odot$ denotes the Hamilton product.

The R-T GM constructs cross-modal long-range dependencies via the cross-attention mechanism. As shown in Fig. \ref{fig3}(b), it first obtains the value and key vectors of the RGB features, denoted as $v_{r}$ and $k_{r}$, respectively:
\begin{equation}
\begin{cases} 
v_{{r}} = Linear(f_{r}) \\
k_{{r}} = Linear(f_{r})
\end{cases} 
\end{equation}
Subsequently, the query vector $q_{t}$ is obtained from $f_{r}$ and {thermal features $f_{t}$ }with an average pooling operation to reduce the computational burden:
\begin{equation}
q_{{t}} = Linear(Pool(Cat(f_{t}, f_{r})))
\end{equation}
where $Pool$ downsamples the feature maps to $7 \times 7$. Finally, long-range dependencies between RGB-T features are captured via self-attention, yielding the global RGB-T features $g_\text{rt}$:
\begin{equation}
g_{\text{rt}} = SoftMax(q_{t}^{\mathrm{T}} \otimes k_{r})\otimes v_{r}
\end{equation} 
where $SoftMax$ and $\otimes$ denote the SoftMax operation and matrix multiplication, respectively.

\begin{figure*}[t]
\centering
\includegraphics[width=\textwidth]{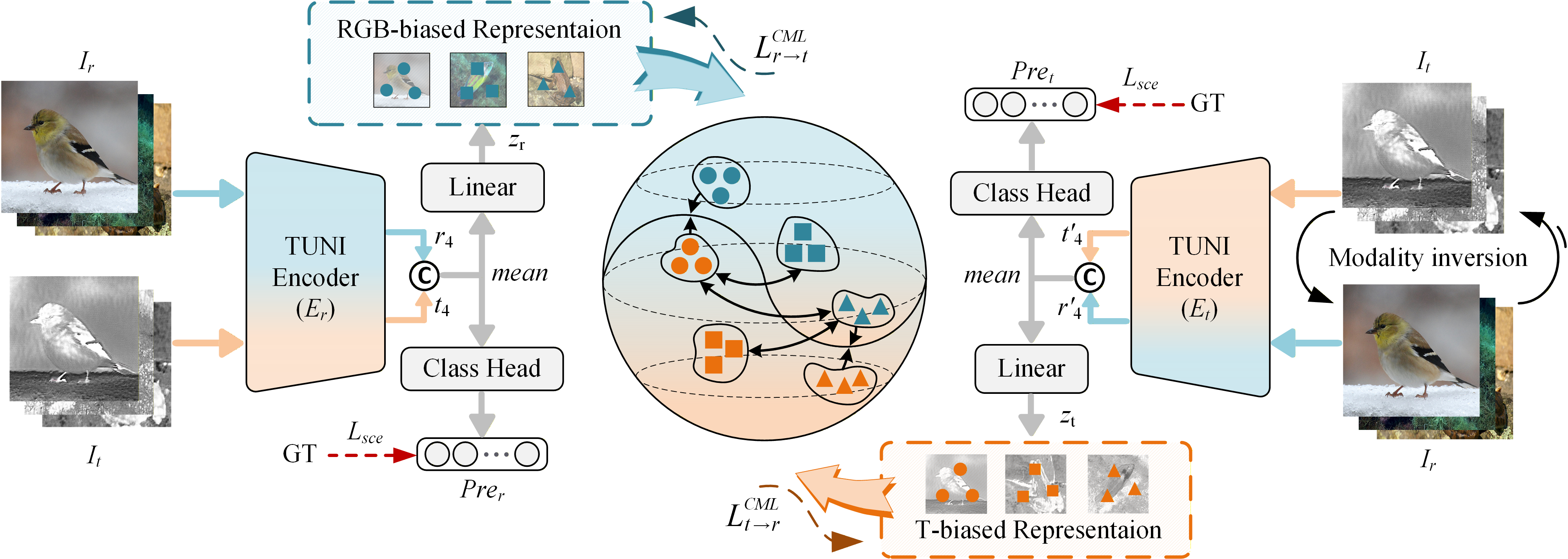}
\caption{Illustration of the MI-CML pre-training. During the pre-training, MI-CML constructs an RGB-dominated encoder $E_r$ and a thermal-dominated encoder $E_t$, then transfers the knowledge between two modality-inverted TUNI encoders with contrastive mutual learning, thereby alleviating the modality bias issue. The pre-trained weights of $E_r$ are retained for the fine-tuning phase.}
\label{fig4}
\end{figure*}

\textit{\textbf{R-T LM:}} In addition to capturing long-range dependencies between RGB-T features, local feature fusion plays a crucial role in leveraging cross-modal information. {In DFormer, cross-modal local feature fusion is achieved using the Hamilton product, taking advantage of the fact that RGB and depth maps exhibit identical geometric structures. However, this method is insufficient for RGB-T feature fusion due to their significant property differences. We further take into account both the consistent and distinct information in RGB-T features, and propose the R-T LM that employs dynamic cosine similarity to select the most informative features, as shown in Fig. \ref{fig3}(c).} Specifically, we extract cross-modal consistent and distinct features, $f_{c}$ and $f_{d}$, using the Hamilton product and absolute difference, respectively. This can be formulated as:
\begin{equation}
\begin{cases}
f_{{c}} = Linear(f_{r})\odot Linear(f_{t}) \\
f_{{d}} = \left|Linear(f_{r})-Linear(f_{t})\right|
\end{cases} 
\end{equation}
Subsequently, two depthwise separable convolutions are applied to further extract the local information from the two feature components:
\begin{equation}
f_{{cd}} = Cat(DWConv(f_{c}), DWConv(f_{d}))
\end{equation}
With the local information extracted, the module needs to select the most informative features of $f_{cd}$. The Squeeze-and-Excitation (SE) block \cite{Squeeze}, a widely adopted channel attention mechanism, provides an effective strategy for this purpose. However, obtaining adaptive weights by pooling along the spatial dimension inevitably results in local information loss. In contrast, we propose an approach that derives adaptive weights by measuring the cosine similarity between the original features and the attention map produced by channel-wise average pooling. {The cosine similarity measures the directional alignment between RGB-T feature representations, capturing modality-invariant semantic consistency while being insensitive to scale differences.} It can be formulated as:
\begin{equation}
W=Cosine(f_{cd},Avg(f_{cd}))
\end{equation}
where $Avg$ denotes the channel-wise average pooling that obtains the attention maps $\in \mathbb{R}^{H \times W \times 1}$; $Cosine$ represents the cosine similarity computation that derives the adaptive weight $ W \in \mathbb{R}^{1 \times 1 \times C}$. Finally, the RGB-T local feature $l_\text{rt}$ is obtained:
\begin{equation}
l_\text{rt}=Linear(f_{cd}\odot SE(W))
\end{equation}
where $SE(.)$ denotes the SE block. With $l_{rr}$, $g_{rt}$, $l_{rt}$, and residual connection, the RGB-T encoder block obtains updated RGB and thermal features $f_{r}'$ and $f_{t}'$:
\begin{equation}
\begin{cases}
f_{r}'=Linear(Cat(l_{rr}, g_{rt}, l_{rt}))+f_{r}\\
f_{t}'=Linear(Cat(l_{rr}, g_{rt}, l_{rt}))+f_{t}
\end{cases}
\end{equation}

\subsection{Pre-training on ImageNet-1K}

ImageNet-1K \cite{ImageNet}, a large-scale classification dataset, is commonly used to pre-train encoders for semantic segmentation models. For TUNI, the main challenge lies in leveraging the RGB-only ImageNet-1K dataset to pre-train a multi-modal encoder. {In RGB-D tasks, DFormer employs a depth estimation network to generate paired RGB and depth images, enabling fully supervised classification pre-training. A similar strategy can be applied to RGB-T tasks by using an RGB-T image translation network\cite{translation} to generate paired RGB-T images, as illustrated in Fig. \ref{fig2}. However, this approach introduces a modality bias issue, where the network tends to over-rely on the RGB branch while underutilizing the thermal branch. This bias is likely caused by the inter-modal parameter imbalance within the TUNI encoder, as the thermal branch contains only half the parameters of the RGB branch. To address this issue, we propose MI-CML, which combines mutual learning with contrastive learning to mitigate modality bias during pre-training.}

\textit {\textbf{MI-CML:}} {Mutual learning\cite{dml} allows models to exchange and exploit the divergencies in their output distributions, thereby enhancing the capabilities of each model. However, previous studies primarily focus on harnessing the complementary strengths of models with different architectures \cite{RCNet, RCL}, leaving cross-modal knowledge exchange largely unexplored. To bridge this gap, MI-CML incorporates mutual learning into multi-modal pre-training, consisting of two stages: modality inversion and contrastive mutual learning.} As shown in Fig. \ref{fig4}, the modality inversion stage constructs two encoders, $E_{r}$ and $E_{t}$, which are dominated by RGB and thermal modalities, respectively. In $E_{r}$, the RGB branch contains twice as many parameters as the thermal branch, whereas in $E_{t}$ the thermal branch has twice as many as the RGB branch. Given paired RGB and pseudo-thermal inputs $I_{r}$ and $I_{t}$, the two encoders extract RGB-dominated and thermal-dominated features, respectively. These features are then projected into their corresponding low-dimensional representations, $Z_{r}$ and $Z_{t}$, which can be formulated as:

\begin{equation}
\begin{cases}
Z_{{r}} = Norm_2(Linear(E_{r}(I_{r}))) \\
Z_{{t}} = Norm_2(Linear(E_{t}(I_{t})))
\end{cases} 
\end{equation}
where $Norm_2$ denotes the $\ell_{2}$ normalization operation along the channel dimension. The concatenation and mean operations are omitted for simplicity. Assuming no modality bias, $Z_{r}$ and $Z_{t}$ are expected to be consistent. Therefore, we introduce contrastive mutual learning to regularize the two representation spaces. For a batch that contains $N$ images, the modality-biased representations are denoted by $\left\{ {Z}^{{i}}_{r} \right\}^{N}_{{i}=1}$ and $\left\{ {Z}^{{i}}_{t} \right\}^{N}_{{i}=1}$. We first fix $E_{t}$ and optimize $E_{r}$: taking $Z_{r}^{i}$ as the anchor, $Z_{t}^{i}$ as the positive sample and {${\{ {Z}^{{j} \neq {i}}_{t} \}}^{N}_{j=1}$ }as negative samples. Note that we do not adopt supervised contrastive learning \cite{SCL} for selecting positive and negative samples, since the Mixup \cite{mixup} used during data augmentation effectively treats each mixed sample as a new class. The contrastive loss function is formulated as:

\begin{equation}
\mathcal{L}_{r\\\to\\t}^{CML}
 = -\sum_{i=1}^{N}\log \frac{\exp(Z^{i}_{r} \cdot Z^{i}_{t} / \tau)}{\sum_{j=1, {j} \neq {i}}^{N}\exp(Z^{i}_{r} \cdot Z^{j}_{t} / \tau)}
\end{equation}
where the temperature coefficient $\tau$ is set to 0.07 following \cite{MCL, SCL}. By pulling RGB-biased representations toward same-class thermal-biased representations and pushing them away from different-class ones, $E_{r}$ learns the thermal-biased feature distribution from $E_{t}$, thereby alleviating the RGB-bias issue. Similarly, we then fix $E_{r}$ and optimize $E_{t}$:
\begin{equation}
\mathcal{L}_{{t}\\\to\\{r}}^{CML}
 = -\sum_{i=1}^{N}\log \frac{\exp(Z^{i}_{t} \cdot Z^{i}_{r} / \tau)}{\sum_{j=1, {j} \neq {i}}^{N}\exp(Z^{i}_{t} \cdot Z^{j}_{r} / \tau)}
\end{equation}

\begin{algorithm}[t]
\caption{{Pseudocode of the MI-CML.}}
\label{alg1}
Input: $D(I_r,I_t,GT)$ \\
Output: $E_r$ \\
\textbf{Step1: Modality Inversion Initialization}

\begin{algorithmic}
\STATE $E_r \gets \text{initialize RGB-dominated TUNI encoder}$
\STATE $E_t \gets \text{initialize thermal-dominated TUNI encoder}$
\STATE $\text{Proj}_r,\text{Proj}_t \gets \text{projection head}$
\STATE $\text{Cls}_r,\text{Cls}_t \gets \text{classification head}$
\end{algorithmic}

\textbf{Step2: Contrastive Mutual Learning Pre-training}

\begin{algorithmic}
\STATE $D_{\text{train}} \gets \text{TrainLoader}(D, \text{batch\_size})$
\STATE for $(I_r, I_t, GT)$ in $\text{enumerate}(D_{\text{train}})$:
\STATE \hspace{0.25cm} \textbf{// optimize $E_r$ while freezing $E_t$}
\STATE \hspace{0.25cm} $E_t.\text{eval}()$
\STATE \hspace{0.25cm} $E_r.\text{train}()$
\STATE \hspace{0.25cm} $f_r,f_t \gets E_r(I_r, I_t),E_t(I_t, I_r)$
\STATE \hspace{0.25cm} $Z_r,Z_t \gets \text{Norm}_2(\text{Proj}_r(f_r)),\text{Norm}_2(\text{Proj}_t(f_t))$
\STATE \hspace{0.25cm} $Pre_r \gets \text{Cls}_r(\mathbf{f}_r)$
\STATE \hspace{0.25cm} $L_r^{class} \gets L_{sce}(Pre_r, GT)$
\STATE \hspace{0.25cm} $L_{r\rightarrow t}^{CML} \gets \text{contrastive loss}(Z_r, Z_t)$
\STATE \hspace{0.25cm} $L_{E_r} \gets \lambda L_r^{class} + (1-\lambda)L_{r\rightarrow t}^{CML}$
\STATE \hspace{0.25cm} update $E_r$

\STATE \hspace{0.25cm} \textbf{// optimize $E_t$ while freezing $E_r$}
\STATE \hspace{0.25cm} $E_r.\text{eval}()$
\STATE \hspace{0.25cm} $E_t.\text{train}()$
\STATE \hspace{0.25cm} $f_r,f_t \gets E_r(I_r, I_t),E_t(I_t, I_r)$
\STATE \hspace{0.25cm} $Z_r,Z_t \gets \text{Norm}_2(\text{Proj}_r(f_r)),\text{Norm}_2(\text{Proj}_t(f_t))$
\STATE \hspace{0.25cm} $Pre_t \gets \text{Cls}_t(\mathbf{f}_t)$
\STATE \hspace{0.25cm} $L_t^{class} \gets L_{sce}(Pre_t, GT)$
\STATE \hspace{0.25cm} $L_{t\rightarrow r}^{CML} \gets \text{contrastive loss}(Z_t, Z_r)$
\STATE \hspace{0.25cm} $L_{E_t} \gets \beta L_t^{class} + (1-\beta)L_{t\rightarrow r}^{CML}$
\STATE \hspace{0.25cm} update $E_t$
\STATE \textbf{return} \boldmath $E_r$ \unboldmath
\end{algorithmic}
\end{algorithm}

This step enables $E_{t}$ to learn from $E_{r}$, preventing the negative effects that may occur if $E_{t}$ falls significantly behind $E_{r}$ as training progresses. For the classification loss, we follow DFormer and adopt a fully supervised loss:
\begin{equation}
\begin{cases}
\mathcal{L}_{r}^{class}= \mathcal{L}_{sce}(Pre_{r}, GT)\\
\mathcal{L}_{t}^{class}= \mathcal{L}_{sce}(Pre_{t}, GT)
\end{cases} 
\end{equation}
where $Pre_{r}$, $Pre_{t}$, and $GT$ denote the classification logit predictions from $E_{r}$ and $E_{t}$, and the ground truth (GT), respectively. $\mathcal{L}_{sce}$ denotes the Soft Cross-Entropy loss. The overall loss function for pre-training $E_{r}$ is defined as:
\begin{equation}
\mathcal{L}_{E_r}=\lambda\mathcal{L}_{r}^{class}+(1-\lambda)\mathcal{L}_{r\\\to\\t}^{CML}
\end{equation}
The overall loss function for pre-training $E_{t}$ is defined as:
\begin{equation}
\mathcal{L}_{E_t}=\beta\mathcal{L}_{t}^{class}+(1-\beta)\mathcal{L}_{t\\\to\\r}^{CML}
\end{equation}
where $\lambda$ and $\beta$ are both set to 0.9. {This alternating update strategy mitigates gradient interference between the two modality-dominated encoders, thereby improving training stability and enabling more reliable cross-modal knowledge transfer. More importantly, it explicitly counteracts the optimization bias caused by asymmetric modality capacity by enforcing alignment between complementary modality-dominated representations. In this way, $E_{r}$ and $E_{t}$ are progressively driven toward a modality-balanced optimum. Since RGB images contain richer information than thermal images, we use the pre-trained $E_{r}$ as the encoder in the subsequent fine-tuning phase. The pseudocode of the MI-CML is provided in Algorithm \ref{alg1}.}

\subsection{Fine-tuning on RGB-T Segmentation Dataset}
Since the TUNI encoder performs both multi-modal feature extraction and cross-modal feature fusion, fine-tuning on RGB-T semantic segmentation datasets is simplified to adding only the segmentation head. {In DFormer, it is assumed that multi-modal fusion has already been achieved within the encoder, and thus only the RGB-branch features are fed into the segmentation head. However, this approach ignores the residual information retained in the thermal branch. A straightforward modification is to concatenate the features from both the RGB and thermal branches for the MLP decoder, yet it yields no noticeable performance improvement. This may be because the residual information in the thermal branch is overwhelmed by the overall features, contributing minimally during network optimization.} {To address this, we propose the MRL decoder, which explicitly focuses on regions where modality-specific predictions are both correct yet different, enabling the model to extract informative residual thermal cues in a more targeted manner.}

\begin{figure}[t!]
      \centering
      \includegraphics[width=0.9\columnwidth]{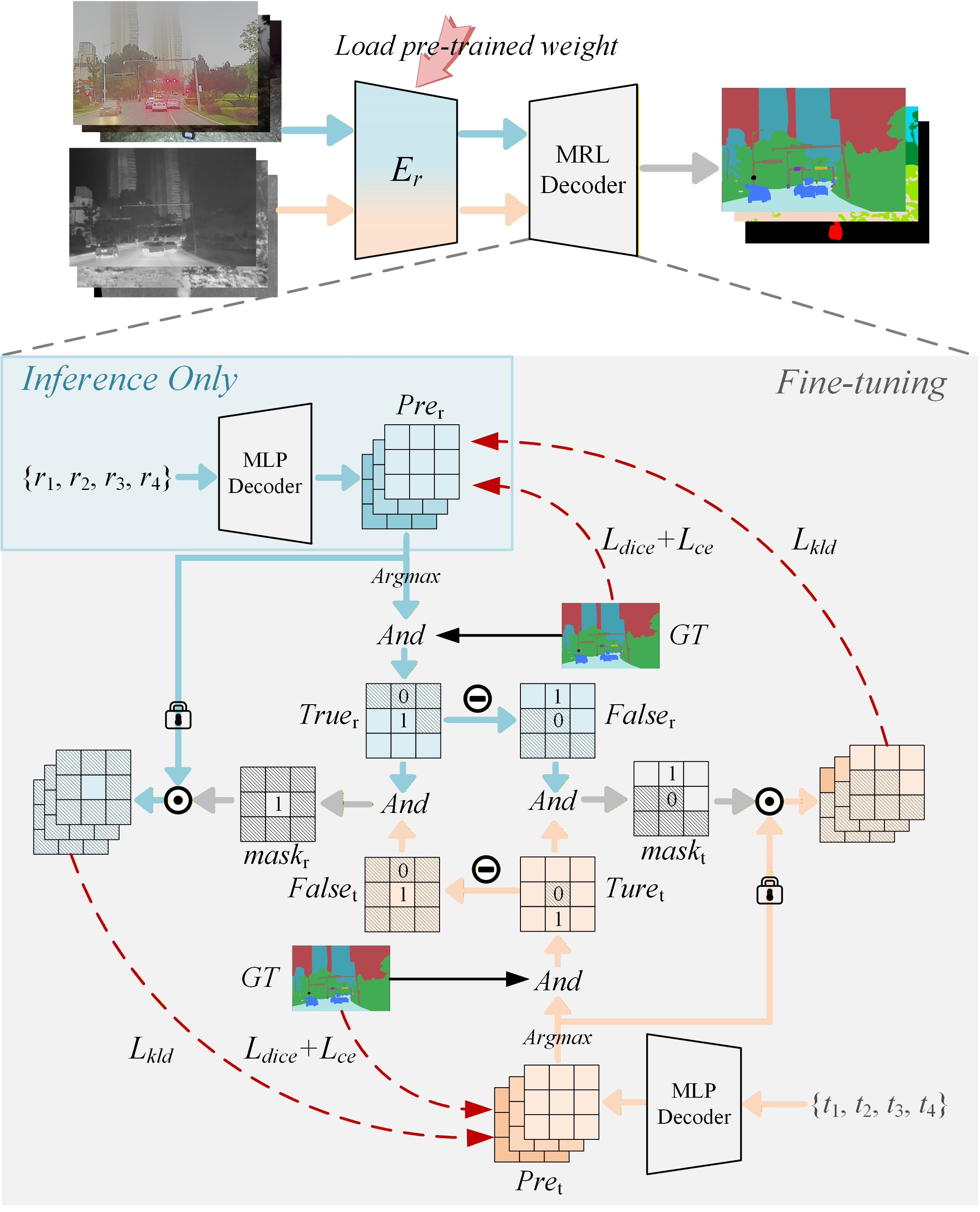}
      \caption{Illustration of MRL decoder. During the fine-tuning phase, MRL focuses on transferring knowledge from regions where the two modality branches both predict correctly but differ, thereby leveraging the residual information in the thermal branch. During inference, only the decoder of the RGB branch is retained, without introducing any additional parameters and inference burden.}
      \label{fig5}
\end{figure}

\textit {\textbf{MRL decoder:}} As shown in Fig. \ref{fig5}, the MRL decoder consists of two MLP decoders that separately decode the information from the RGB and thermal branches, as well as MRL that leverages the complementary strengths of one branch to rectify the other. The four-level RGB and thermal features extracted by the TUNI encoder from the input images are denoted as $\{r_1, r_2, r_3, r_4\}$ and $\{t_1, t_2, t_3, t_4\}$, and the two MLP decoders are denoted as $D_r$ and $D_t$, respectively. After decoding the features from the two branches, the logit segmentation predictions $Pre_{r}$ and $Pre_{t}$ are obtained:
\begin{equation}
\begin{cases}
Pre_{r}=D_r(r_1,r_2,r_3,r_4)\\
Pre_{t}=D_t(t_1,t_2,t_3,t_4)
\end{cases}
\end{equation}

If $Pre_{r}$ and $Pre_{t}$ learn from each other directly, the effect is almost negligible. This is likely because the information from the two branches is already highly entangled, causing the informative divergencies that are worth learning to be overwhelmed. {Instead, we restrict mutual learning to regions where both modalities make correct yet different predictions. These regions correspond to reliable samples with complementary modality-specific cues, and thus provide more informative supervision for cross-modal knowledge transfer, particularly for recovering residual thermal information that is suppressed during feature fusion.}
Specifically, we leverage the GT to identify the correct and incorrect prediction regions in $Pre_{r}$ and $Pre_{t}$:
\begin{equation}
\begin{cases}
True_r=GT \wedge Argmax(Pre_{r})\\
False_r=1-True_r
\end{cases}
\end{equation}
\begin{equation}
\begin{cases}
True_t=GT \wedge Argmax(Pre_{t})\\
False_t=1-True_t
\end{cases}
\end{equation}
where $Argmax$ denotes the argmax operation along the channel dimension to obtain semantic maps, and $\wedge$ denotes the logical AND operation. {$True_r$ and $True_t$ denote the correctly predicted regions of the RGB and thermal decoders, respectively, while $False_r$ and $False_t$ correspond to the regions where the predictions are incorrect.} For the RGB branch, we select regions where the RGB branch predicts incorrectly while the thermal branch predicts correctly as the learning targets, and vice versa for the thermal branch:

\begin{equation}
\begin{cases}
mask_t=True_t \wedge False_r\\
mask_r=True_r \wedge False_t
\end{cases}
\end{equation}
In this way, we obtain the learning region mask $mask_t$ and $mask_r$, which focus the mutual learning of the two branches on regions that are both correct and divergent. The Kullback–Leibler (KL) divergence is then employed as the mutual learning loss function:

\begin{equation}
\begin{cases}
\mathcal{L}_{r \to t}^{MRL}=\mathcal{L}_{KLD}(Pre_{r}, mask_t(Pre_{t}))\\
\mathcal{L}_{t \to r}^{MRL}=\mathcal{L}_{KLD}(Pre_{t}, mask_r(Pre_{r}))
\end{cases}
\end{equation}
Following \cite{CM-SSM}, we employ the Dice loss and Cross-Entropy loss for semantic segmentation. The loss function for $D_r$ and $D_t$ are defined as follows:
\begin{equation}
\begin{cases}
\mathcal{L}_{D_r}=\mathcal{L}_{ce}(Pre_{r},GT)+
\mathcal{L}_{dice}(Pre_{r},GT)+
\mathcal{L}_{r \to t}^{MRL}\\
\mathcal{L}_{D_t}=\mathcal{L}_{ce}(Pre_{t},GT)+
\mathcal{L}_{dice}(Pre_{t},GT)+
\mathcal{L}_{t \to r}^{MRL}
\end{cases}
\end{equation}
Finally we obtain the total loss function for the fine-tuning phase:
\begin{equation}
\mathcal{L}_{seg}=\gamma\mathcal{L}_{D_r}+(1-\gamma)\mathcal{L}_{D_t}
\end{equation}
where $\gamma$ is set to 0.8, allowing the RGB branch to play a dominant role during optimization. {During the inference stage, only $D_r$ is retained because the cross-modal knowledge has been implicitly transferred to the RGB branch, ensuring that the MRL decoder introduces no additional inference overhead and preserving the model’s lightweight design and real-time performance. In this way, MRL further enhances thermal information utilization during the fine-tuning stage and completes the overall TUNI framework.}

\section{Experimental Results}

\subsection{Dataset and Evaluation Metrics}

We pre-train our models on ImageNet-1K \cite{ImageNet} and employ five public RGB-T datasets for the fine-tuning and evaluating of our methods.

\textbf{The MSRS dataset\cite{MSRS}} contains 1,444 pairs of aligned RGB-T images under urban scenes, including 8 sub-classes: car, person, bike, curve, car stop, guardrail, color cone, and  bump. It removes misaligned image pairs from MFNet\cite{MFNet}, and collects 715 daytime and 729 nighttime image pairs. An image enhancement algorithm based on dark channel prior is leveraged to optimize the contrast and signal-to-noise of thermal images. All images have a spatial resolution of 640 × 480 pixels and are divided into train/test sets at a ratio of 8:2.

\textbf{The FMB dataset  \cite{FMB}} comprises 1500 pairs of RGB-T images under different illumination conditions in urban scenes. It labels 14 categories, including road, sidewalk, building, traffic light, traffic sign, vegetation, sky, person, car, truck, bus, motorcycle, bicycle and pole. All images have a spatial resolution of 800 × 600 pixels. The dataset is divided into train/test sets at a ratio of 8:2.

\textbf{The PST900 dataset  \cite{PST900}} comprises 894 pairs of RGB-T images captured in challenging underground environments. It provides semantic segmentation annotations for 4 classes: hand drill, backpack, fire extinguisher and survivor. Both the RGB and thermal images have a spatial resolution of 1280 × 720 pixels. The dataset is divided into train/test sets at a ratio of 2:1.

\textbf{The CART dataset  \cite{CART}} contains 2282 pairs of aligned RGB-T images captured in various terrains, including rivers, lakes, coastlines, deserts and forests. It provides semantic segmentation labels for 10 classes: bare ground, rocky terrain, developed structures, road, shrubs, trees, sky, water, vehicles, and person. Both the RGB and thermal images have a spatial resolution of 960 × 640 pixels. The dataset was randomly partitioned into train/val/test sets at a 6:1:1 ratio.

\textbf{The SUS dataset  \cite{SUS}} comprises 1035 pairs of aligned RGB-T images collected in snowy urban scenes. It provides semantic annotations for five classes: road, sidewalk, person, car, and bicycle. All images have a spatial resolution of 640 × 480 pixels. The dataset is divided into train/val/test sets at a ratio of 7:1:2.

We utilize the intersection over union (IoU) and mean intersection over union (mIoU) to assess sub-class and overall segmentation performance, respectively. The floating-point operations (FLOPs), the number of parameters (Params), and frames per second (FPS) are used to evaluate the model complexity. {FLOPs are measured at an input resolution of 640 $\times$ 480, and FPS is evaluated with a batch size of 1, including only the model inference time and excluding preprocessing and postprocessing steps.}

\begin{table}[t!]
\caption{Detailed configuration of various versions of TUNI encoder. $C$=($C_{rgb}$, $C_{t}$), where $C_{rgb}$ and $C_{t}$ denote the channel numbers of the RGB and thermal branches, respectively. $Depth$ denotes the number of RGB-T encoder blocks in each stage. The
Params and FLOPS are measured at an image resolution of 640 × 480.}
\centering
\begin{tabular}{c|c|ccc}
\toprule
\rowcolor[gray]{.9}
Stage&	Output Size&	TUNI-T& TUNI-S & TUNI-B\\ 
\midrule
Stem&	H/4×W/4&	C=(16, 8)&	(24, 12)&(32, 16)\\
1&	H/4×W/4&	C=(32, 16)&	(48, 24)&(64, 32)\\
2&	H/8×W/8&	C=(64, 32)&	(96, 48)&(128, 64)\\
3&	H/16×W/16&	C=(128, 64)&	(192, 96)&(256, 128)\\
4&	H/32×W/32&	C=(256, 128)&	(384, 192)&(512, 256)\\
\midrule
Depths&-&(2, 2, 4, 2)&(2, 2, 4, 2)&(3, 3, 12, 2)\\
Params&-&4.60&10.18&29.06\\
FLOPs&-&5.36&11.57&36.91\\
\bottomrule
\end{tabular}
\label{table1}
\end{table}

\subsection{Implementation Details}

\textit{\textbf{1) TUNI Encoder:}} We develop three versions of the TUNI encoder--TUNI-T, TUNI-S and TUNI-B--targeting real-time, balanced and high-performance RGB-T semantic segmentation models. The output size, channel dimension and depth of each stage, along with the corresponding Params and FLOPs, are summarized in Table \ref{table1}.

\textit{\textbf{2) Pre-training:}} The pre-training is conducted on ImageNet-1K with 8 NVIDIA H20 GPUs. An RGB-T translation model \cite{translation} is utilized to generate aligned RGB-T image pairs and a classifier head is added to build the classification model. The RGB-T image pairs are resized to 224 × 224. The data augmentation strategies related to color are only used for RGB images, while other common strategies are simultaneously performed on RGB and thermal images, e.g., random rotation, mixup, cutmix. During the training, $E_r$ and $E_t$ are optimized sequentially using two AdamW optimizers with the initial learning rate of 1e-3 and weight decay of 5e-2, and the batch size is set to 1024. When one encoder is being trained, the other is frozen to ensure training stability. We train the model for 300 epochs and the checkpoint of $E_r$ that achieves the best validation performance is retained.

\textit{\textbf{3) Fine-tuning:}} The fine-tuning is conducted on MSRS, FMB, PST900, CART and SUS on an RTX 4090 GPU. Before training, the TUNI encoder is initialized with the pre-trained weights obtained from the pre-training phase, whereas the segmentation head is initialized randomly. Image resizing, random cropping and flipping were used to augment the data during training, as configured in the original benchmarks \cite{FMB, PST900, CART, MSRS, SUS}. The training is optimized using Ranger with the weight decay of 5e-4 and initial learning rate of 1e-4, multiplied by $(1-\frac{iter}{max\_iter})^{power}$ during training, with a power of 0.9. For MSRS, FMB, PST900, CART, and SUS, the training epochs are set to 500, 80, 200, 300 and 200, with corresponding batch sizes of 4, 4, 4, 8, and 4, respectively. {All results are averaged over three independent runs to ensure stable evaluation.}

\subsection{Comparison with SOTA Methods}

\textit{\textbf{1) Comparison Methods.}} On MSRS, FMB,  PST900, CART and SUS, we compare TUNI with 15 SOTA methods, including: GMNet\cite{GMNet}, CMX\cite{CMX}, SGFNet\cite{SGFNet}, CMNext\cite{CMNext}, CLNet-T\cite{CLNet}, DFormer\cite{DFormer}, MiLNet\cite{MILNet}, AGFNet\cite{AGFNet}, DFormerV2\cite{DFormerV2}, CPAL\cite{CPAL}, MCNet-T\cite{SUS}, Sigma\cite{sigma}, CM-SSM\cite{CM-SSM}, FGDNet-S\cite{FGDNet-S}, TUNI$^{\dagger}$\cite{TUNI}.

\begin{table*}[t]
\caption{Complexity and performance comparison with SOTA methods on MSRS, FMB, PST900, CART and SUS. TUNI$^{\dagger}$ denotes the conference version of our framework. The
Params, FLOPS, and FPS are measured at an image resolution of 640 × 480. ‘–’ denotes missing information and non-public models, and `\ding{55}' denotes incompatibility with the platform. ‘Jetson’ denotes the Jetson Orin NX. For the Jetson platform, numbers outside parentheses report CUDA-based inference results, whereas numbers in parentheses correspond to TensorRT inference with BF16 precision. The best and second-best results in each column are highlighted in \textbf{bold} and \underline{underline}, respectively.}
\centering
\label{table2}
\renewcommand{\arraystretch}{1.25}
\begin{tabular}{cccccccccccc}
\toprule
\multirow{2.5}{*}{Model} & \multirow{2.5}{*}{Publication} &\multirow{2.5}{*}{Backbone}  &\multirow{2.5}{*}{Params(M)↓}  &\multirow{2.5}{*}{FLOPs(G)↓}  & \multicolumn{2}{c}{FPS↑} & \multicolumn{5}{c}{mIoU↑} \\
\cmidrule(lr){6-7} \cmidrule(lr){8-12}
& & & && RTX 4090 & Jetson & MSRS & FMB & PST900 & CART & SUS \\
\midrule
GMNet & TIP’21 & ResNet101 & 191.24 & 195.37 & 41 & 4 (\ding{55}) & 73.9 & 49.2 & 84.1 & 72.7 & 81.2 \\
CMX & TITS’23 & Mit-B2 & 66.57 & 67.20 & 63 & 3 (13) & 75.3 & 61.1 & 84.9 & 74.0 & 81.2 \\
SGFNet & TCSVT’23 & ResNet50 & 125.25 & 144.83 & 42 & 2 (\ding{55}) & 75.9 & 56.0 & 82.8 & 68.5 & 79.8 \\
CMNext & ICCV’23 & Mit-B2 & 58.68 & 68.70 & 67 & 3 (11) & 74.8 & 59.0 & 83.9 & 72.1 & 80.0 \\
CLNet-T & KBS’24 & Mit-B4 & 130.84 & 217.85 & 28 & 1 (\ding{55}) & 76.3 & 60.4 & 80.8 & 73.7 & 80.9 \\
DFormer & ICLR’24 & DFormer-B & 29.5 & 41.9 & 69 & 5 (15) & 77.6 & 61.2 & 85.4 & 73.9 & 81.1 \\
MiLNet & TIP’25 & Mit-B3 & 92.29 & 136.32 & 25 & 2 (\ding{55}) & 74.7 & 61.8 & 85.1 & 74.6 & 81.0 \\
AGFNet & TITS’25 & ResNet50 & 72.19 & 219.11 & - & - & - & 60.5 & 84.8 & - & - \\
DFormerV2 & CVPR’25 & DFormerV2-S & 26.70 & 33.90 & 33 & 2 (11) & 74.1 & 61.5 & 83.9 & 72.4 & 80.3 \\
CPAL & TCSVT’25 & Swin-T & 52.70 & - & - & - & - & 60.9 & 86.8 & - & - \\
MCNet-T & TITS’25 & ConvNextV2-B & 199.79 & 278.85 & 32 & \ding{55} & 73.8 & \underline{63.6} & 84.6 & 73.7 & \underline{83.5} \\
Sigma & WACV’25 & Vmamba-T & 48.3 & 89.5 & 19 & \ding{55} & 78.9 & 61.8 & \underline{88.6} & 74.0 & 82.4 \\
CM-SSM & IROS’25 & EfficientVit-B1 & \underline{10.34} & \underline{12.59} & 114 & 9 (\textbf{42}) & 77.4 & 59.0 & 85.9 & 74.6 & 82.0 \\
FGDNet-S & IF’25 & Res18+Mit-B0 & 17.96 & 38.71 & - & - & - & 49.7 & 72.2 & - & - \\
TUNI$^{\dagger}$ & ICRA’26 & TUNI-S & 10.63 & 17.16 & \underline{120} & \underline{11} (27) & 77.3 & 62.4 & 87.3 & 74.7 & 81.2 \\
\midrule
\rowcolor{cyan!10}
\multirow{3}{*}{TUNI} & \multirow{3}{*}{-} & TUNI-T & \textbf{4.99} & \textbf{10.82} & \textbf{135} & \textbf{16} (\underline{35}) & 78.6 & 62.4 & 86.4 & 73.5 & 82.1 \\
\rowcolor{cyan!10}
TUNI (Ours)&-& TUNI-S & 10.63 & 17.16 & \underline{120} & \underline{11} (27) & \underline{79.7} & 63.5 & 87.3 & \underline{75.5} & 82.8 \\
\rowcolor{cyan!10}
& & TUNI-B & 29.57 & 42.64 & 60 & 6 (14) & \textbf{80.7} & \textbf{66.3} & \textbf{89.1} & \textbf{75.7} & \textbf{83.9} \\
\bottomrule
\end{tabular}
\end{table*}

\textit{\textbf{2) Quantitative Comparison.}} Table \ref{table2} summarizes the complexity and performance comparisons with SOTA models across the five datasets, and Table \ref{table3} presents the IoU results for the sub-classes in each dataset. 

\textit{TUNI-B:} As shown in Table \ref{table2}, TUNI-B achieves the best mIoU across all five datasets. It surpasses the best SOTA models by 1.6\% (MSRS), 2.7\% (FMB), 0.5\% (PST900), 1.0\% (CART) and 0.4\% (SUS), demonstrating robust generalization capability. Compared with MCNet-T, TUNI-B attains better segmentation accuracy with only 1/7 of the Params and FLOPs, indicating a significantly improved efficiency–performance trade-off. When compared with Sigma, TUNI-B achieves 3× faster inference speed and remains fully compatible with edge deployment, making it more suitable for real-time perception. Moreover, TUNI-B outperforms DFormer, which has a comparable model size, by 3.1\% (MSRS), 5.1\% (FMB), 3.7\% (PST900), 1.8\% (CART), and 2.8\% (SUS). {This consistent improvement across diverse datasets highlights the strong generalization ability of TUNI, which benefits from its unified design that effectively leverages thermal information during both pre-training and fine-tuning stages.}

\textit{TUNI-S:} TUNI-S offers a favorable trade-off between performance and efficiency. Compared with TUNI-B, it sacrifices limited accuracy while halving the model size and doubling the inference speed. While achieving state-of-the-art performance on MSRS, CART, and SUS, it delivers 120 FPS on an RTX 4090 and 27 FPS on a Jetson Orin NX (TensorRT, BF16), demonstrating real-time capability on resource-constrained edge platforms. Compared with TUNI$^{\dagger}$ (conference version) that ignores the modality bias during pre-training and the thermal residual information during fine-tuning, TUNI-S achieves improvements of 2.4\%, 1.1\%, and 1.6\% on MSRS, FMB, and SUS, respectively. The lack of performance gains on PST900 and CART can be explained by their RGB-dominated data distribution, in which the benefits of enhanced thermal information utilization are less evident.

\begin{table*}[t]
\caption{Performance comparison of some sub-classes with SOTA methods on MSRS, FMB, PST900, CART and SUS.}
\centering
\label{table3}
\setlength{\tabcolsep}{4pt}
\renewcommand{\arraystretch}{1.25}
\begin{tabular}{cccccccccccccccc}
\toprule
\multirow{2.5}{*}{Model}&\multicolumn{3}{c}{MSRS} & \multicolumn{3}{c}{FMB}&\multicolumn{3}{c}{PST900}&\multicolumn{3}{c}{CART}&\multicolumn{3}{c}{SUS}\\
\cmidrule(lr){2-4} \cmidrule(lr){5-7}\cmidrule(lr){8-10}\cmidrule(lr){11-13}\cmidrule(lr){14-16}
&	Car&	Person&	Color Cone &Person&	Car&	Pole&	Hand Drill&	Fire Exiti.&	Survivor&	Shrub&	Vehicle&	Person&	Person&	Car&	Bicycle\\
\midrule
Dformer&	90.6&	72.6&	63.3&	66.9&	79.9&	46.2&	75.0&	81.8&	82.2&	\textbf{72.5}&	53.4&	21.5&	74.6&	88.1&	76.8\\
MiLNet&	89.6&	70.6&	56.2&	66.9&	81.7&	48.4&	80.3&	77.2&	78.7&	71.0&	56.3&	30.3&	75.5&	87.9&	75.5\\
Sigma&	90.9&	74.4&	64.8&	69.7&	\underline{81.6}&	\underline{49.5}&	\textbf{81.9}&	\textbf{89.8}&	82.7&	71.1&	56.8&	24.3&	78.5&	88.9&	77.8\\
CM-SSM&	89.4&	74.1&	65.2&	66.1&	79.1&	49.1&	80.5&	85.7&	78.1&	71.3&	\textbf{62.0}&	24.5&	78.9&	88.3&	75.5\\
TUNI$^{\dagger}$&	90.0&	73.9&	66.0&	71.2&	78.7&	47.3&	78.2&	86.1&	81.7&	71.1&	60.3&	27.5&	79.0&	88.7&	76.8\\
\midrule
\rowcolor{cyan!10}
TUNI-T&	88.7&	75.7&	63.8&	70.1&	79.1&	46.2&	78.8&	86.9&	79.3&	70.3&	56.7&	28.1&	79.0&	88.5& 76.8\\
\rowcolor{cyan!10}
TUNI-S&	\textbf{91.7}&	\textbf{77.1}&	\underline{67.2}&	\underline{72.3}&	80.4&	48.2&	79.2&	86.3&	\underline{83.0}&	71.2&	59.2&	\underline{31.7}&	\underline{79.7}&	\underline{89.4}&	\underline{77.9}\\
\rowcolor{cyan!10}
TUNI-B&	\underline{91.2}&	\underline{76.7}&	\textbf{69.1}&	\textbf{73.1}&	\textbf{82.7}&	\textbf{50.7}&	\underline{81.6}&	\underline{89.2}&	\textbf{83.4}&	\underline{71.7}&	\underline{60.4}&	\textbf{32.1}&	\textbf{81.2}&	\textbf{89.9}&	\textbf{79.1}\\
\bottomrule
\end{tabular}
\end{table*}

\begin{figure*}[!t]
\centering
\includegraphics[width=\textwidth]{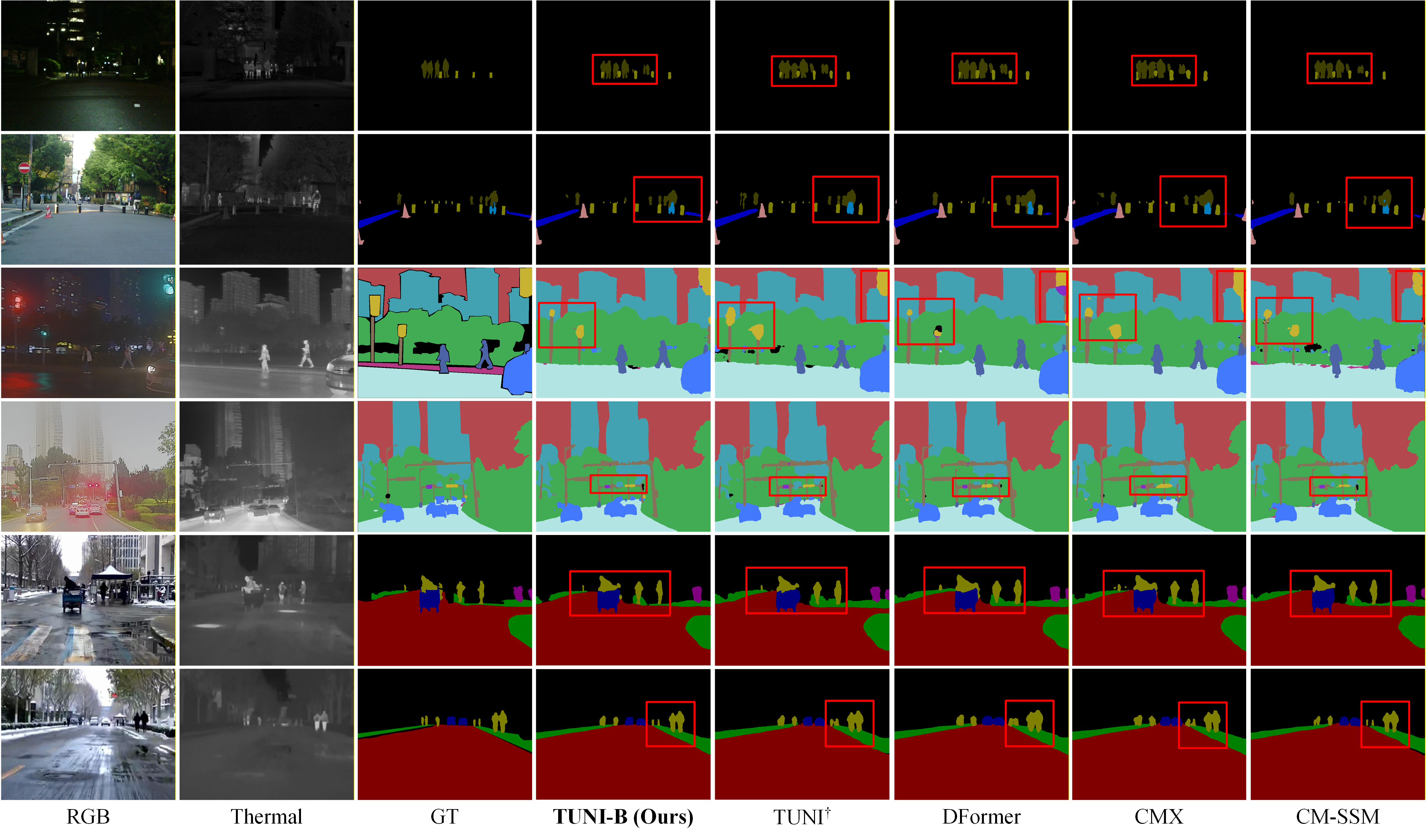}
\caption{Visual comparison of segmentation maps produced by TUNI-B, DFormer, TUNI$^{\dagger}$, CMX, and CM-SSM on MSRS (top two rows), FMB (middle two rows), and SUS (bottom two rows).}
\label{fig6}
\end{figure*}

\textit{TUNI-T:} We provide the TUNI-T for scenarios with stringent computational constraints or high real-time requirements.  It is extremely lightweight, requiring only 4.99M Params and 10.82G FLOPs, and delivers high-speed inference of 135 and 35 FPS on the RTX 4090 and Jetson, respectively, while remaining competitive with SOTA models across diverse datasets. Moreover, TUNI-T, with a smaller model size, outperforms TUNI$^{\dagger}$ on MSRS and SUS while matching its performance on FMB, further validating the effectiveness of the proposed MI-CML pre-training and MRL fine-tuning framework.

\textit{Sub-class Comparison:} As shown in Table \ref{table3}, MiLNet and Sigma are included as high-performance baselines, whereas CM-SSM and TUNI are considered lightweight models. TUNI-B consistently achieves the best or second-best results across all subclasses, and TUNI-S demonstrates superior segmentation accuracy compared with other lightweight models. For categories with pronounced thermal signatures (e.g., `person', `car' and `vehicle'), our framework yields more accurate segmentation results, validating its enhanced utilization of thermal information.

\textit{\textbf{3) Qualitative Evaluation.}} Fig.~\ref{fig6} presents a qualitative comparison among TUNI-B, DFormer, TUNI$^{\dagger}$, CMX, and CM-SSM on MSRS, FMB, and SUS, where the red rectangular boxes highlight representative regions. Overall, the segmentation results produced by TUNI-B are the closest to the ground truth. {In particular, as highlighted in the red boxes, TUNI-B achieves more accurate predictions for objects with prominent thermal characteristics, such as ‘person’ and ‘traffic sign’, especially under low-illumination or glare conditions where other methods fail to capture clear object boundaries.} These observations indicate that our framework achieves more effective utilization of thermal information. Moreover, TUNI-B exhibits improved boundary delineation between classes, which can be attributed to the proposed R–T LM.

\begin{figure*}[t]
\centering
\includegraphics[width=\textwidth]{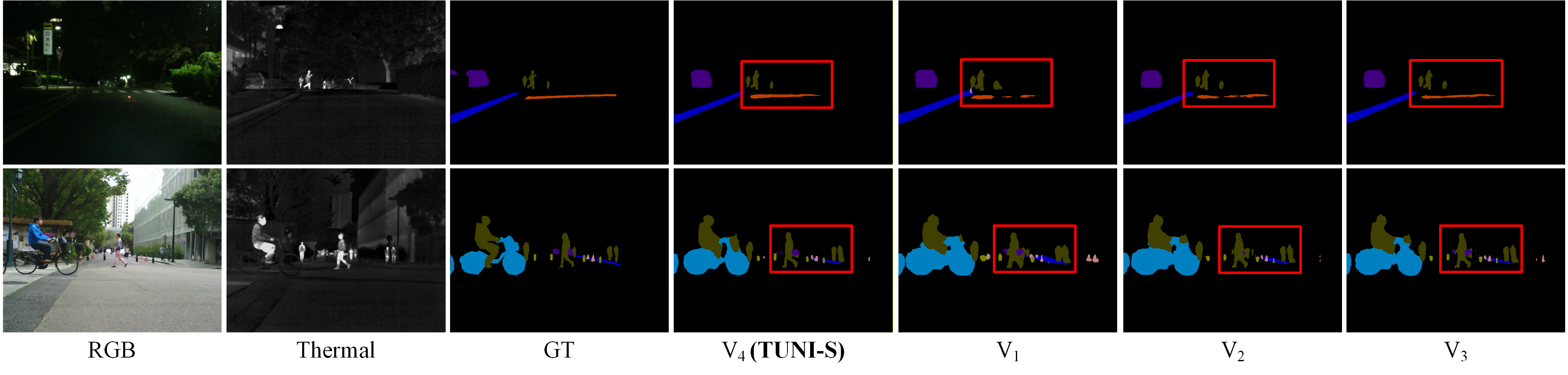}
\caption{Visual comparison of segmentation maps produced by four variations on MSRS. $V_{1}$ represents a baseline that
directly applies the DFormer framework to RGB-T semantic
segmentation; $V_{2}$ introduces the R–T LM to better
exploit local RGB–T features; $V_{3}$ further incorporates
MI-CML to alleviate RGB-bias during pre-training;
$V_{4}$ employs the MRL decoder to leverage thermal residual
information during fine-tuning.}
\label{fig7}
\end{figure*}

\subsection{Ablation Studies}

To validate the effectiveness of each component in the TUNI framework, including the TUNI encoder, MI-CML pre-training, and MRL fine-tuning, we conduct a series of ablation studies.

\textit{\textbf{1) Overall Ablation Study.}} We first conduct an overall ablation study to evaluate the individual components of TUNI on the MSRS and SUS datasets. As shown in Table \ref{table4}, Variant 1 represents a baseline that directly applies the DFormer framework to RGB-T semantic segmentation; Variant 2 introduces the R–T LM to better exploit local RGB–T features; Variant 3 further incorporates MI-CML to alleviate RGB-bias during pre-training; Variant 4 employs the MRL decoder to leverage thermal residual information during fine-tuning. Compared with the baseline, Variants 2, 3, and 4 improve the mIoU by 2.0\%, 3.3\%, and 4.4\% on MSRS, respectively, progressively validating the effectiveness of the proposed methods. Fig.~\ref{fig7} presents the visualization results of the four variants, where the boundaries of categories such as `bump' and `person' become progressively clearer as the proposed modules are introduced. This consistent improvement indicates that the proposed components are coherently designed around enhancing thermal information utilization, forming a unified framework that progressively improves fine-grained segmentation. Additionally, on the SUS dataset, Variants 2, 3, and 4 improve the mIoU by 1.4\%, 0.9\%, and 0.7\%, respectively, further validating the effectiveness of each component on a different dataset. Since MSRS contains high-quality enhanced thermal images, it serves as a more suitable benchmark for evaluating the effectiveness of thermal information utilization. Therefore, all subsequent ablation studies are conducted on the MSRS dataset. {A video demo visually demonstrating the performance improvements over the baseline and ablation baselines is provided\footnote{https://github.com/xiaodonguo/TUNI-v2}.} 

\begin{table}[t]
\caption{{Effectiveness of the key components of TUNI framework.}}
\centering
\begin{tabular}{c|ccc|cc}
\toprule
\rowcolor[gray]{.9}
Variant & R-T LM & MI-CML & MRL & MSRS & SUS\\
\midrule
1       & -       &   -     &  -    & 75.3 & 79.8\\
2       & \ding{52} &     -   &   -   & 77.3 & 81.2\\
3       & \ding{52} & \ding{52} &   -   & 78.6 & 82.1\\
\rowcolor{cyan!10}
4       & \ding{52} & \ding{52} & \ding{52} & \textbf{79.7} &\textbf{82.8}\\
\bottomrule
\end{tabular}
\label{table4}
\end{table}

\begin{table}[t]
\caption{Effectiveness of sub-modules in the TUNI Encoder.}
\centering
\begin{tabular}{c|cc|c}
\toprule
\rowcolor[gray]{.9}
Module&Params&FLOPs&mIoU           \\ 
\midrule
w/o R-R LM  &   9.14&	15.65&	76.3\\
w/o R-T GM  &	9.52&	16.43&	76.0\\
w/o R-T LM  &	9.54&	16.16&	75.0\\
DFormer Encoder& 10.36& 16.96&75.3\\
\rowcolor{cyan!10}
TUNI Encoder&	10.63&	17.16&	\textbf{77.3}\\
\bottomrule
\end{tabular}
\label{table5}
\end{table}
\textit{\textbf{2) Effectiveness of Modules in TUNI Encoder.}} Table \ref{table5} presents the ablation results of various TUNI encoder variants, including complexity and performance indicators. ‘w/o R-R LM’ and ‘w/o R-T GM’ denote the TUNI encoder removing R-R LM and R-T GM, respectively. Compared with TUNI encoder, their performance decreases by 1.0\% and 1.3\% in mIoU, demonstrating the necessity of retaining these two modules. ‘w/o R-T LM’ refers to the TUNI encoder with the R-T LM removed. R-T LM introduces only 1.09M parameters and 1.00G FLOPs, while improving the mIoU by 2.3\%, confirming its efficiency in fusing RGB and thermal local features. ‘DFormer Encoder’ denotes the original DFormer encoder, which differs from TUNI by employing a simple Hamilton product for cross-modal local feature fusion. Its performance is comparable to that of the “w/o R–T LM” variant, indicating that such a simple fusion strategy is ineffective for RGB–T local feature fusion. This further validates the necessity of our proposed R-T LM.

\begin{table}[t!]
\caption{Effectiveness of components in R-T LM.}
\centering
\begin{tabular}{c|ccc|cc|c}
\toprule
\rowcolor[gray]{.9}
Variant& Co\&Di &	SE&	Cosine&	Params&	FLOPs&	mIoU\\ 
\midrule
1& -   &-	&-	&	9.54&	16.16&	75.0\\
2&\ding{52}	&-	&-	&	10.52&	17.16&	75.9\\
3&\ding{52}	&\ding{52}	&-	&10.63&	17.16&	76.3\\
\rowcolor{cyan!10}
4&\ding{52}	&\ding{52}	&\ding{52}&	10.63&	17.16&	\textbf{77.3}\\
\bottomrule
\end{tabular}
\label{table6}
\end{table}

\begin{table}[t!]
\caption{Effect of various pre-training modalities.}
\centering
\begin{tabular}{ccc|c}
\toprule
\rowcolor[gray]{.9}
RGB-RGB     & RGB-D        & RGB-T      & mIoU\\
\midrule
\ding{52}& -& -&75.5\\
-& \ding{52}& -&76.0\\
\rowcolor{cyan!10}
-& -&\ding{52} &\textbf{77.3}\\
\bottomrule
\end{tabular}
\label{table7}
\end{table}

\begin{table}[t!]
\caption{Comparison of various RGB-T pre-training strategies.}
\centering
\begin{tabular}{c|ccc|c}
\toprule
\rowcolor[gray]{.9}
Encoder     &Vanilla        &MCL      &MI-CML   &mIoU\\
\midrule
$E_t$        &\ding{52}       &    -   &  -  &75.6\\
$E_r$     &\ding{52}      &        -   &  -  &77.3\\
$E_r$     &  -    &\ding{52}           &  -  &77.3\\
$E_t$        &   -    &   -    &\ding{52}    &77.0\\
\rowcolor{cyan!10}
$E_r$     &   -   &       -    &\ding{52}    &\textbf{78.6}\\
\bottomrule
\end{tabular}
\label{table8}
\end{table}

\textit{\textbf{3) Effectiveness of Components in R-T LM.}} We conduct a fine-grained ablation study to assess the effectiveness of individual components in R-T LM, with results shown in Table \ref{table6}. ‘Co\&Di’ denotes the interaction of consistent and distinct cross-modal features, including linear and depthwise separable convolution layers. ‘SE’ and ‘Cosine’ denote the squeeze-and-excitation and cosine similarity components, respectively. Incorporating ‘Co\&Di’ yields mIoU improvements of 0.9\%, which demonstrates the effectiveness of multi-attribute feature interaction. However, adding ‘SE’ brings little improvement, likely due to the weights obtained from global spatial averaging lacking local feature representativeness. Without introducing additional computational or parameter overhead, incorporating cosine similarity yields mIoU gains of 1.0\%, which confirms the effectiveness of the proposed approach.

\textit{\textbf{4) Effectiveness of RGB-T Pre-training.}} Table \ref{table7} presents three pre-training strategies: ‘RGB-RGB’ denotes using identical RGB pairs as input to the TUNI encoder, ‘RGB-D’ refers to RGB-D pairs generated as in \cite{DFormer}, and ‘RGB-T’ represents our RGB-T pre-training strategy. Compared with ‘RGB-RGB’, our approach improves mIoU by 1.8\%. This indicates that pre-training with RGB-T data effectively enhances the encoder’s ability to capture features from the thermal modality. Compared with ‘RGB-D’, our approach improves mIoU by 1.3\%, which demonstrates that the modality used during pre-training affects the performance of segmentation.

\begin{figure}[t]
      \centering
      \includegraphics[width=0.8\columnwidth]{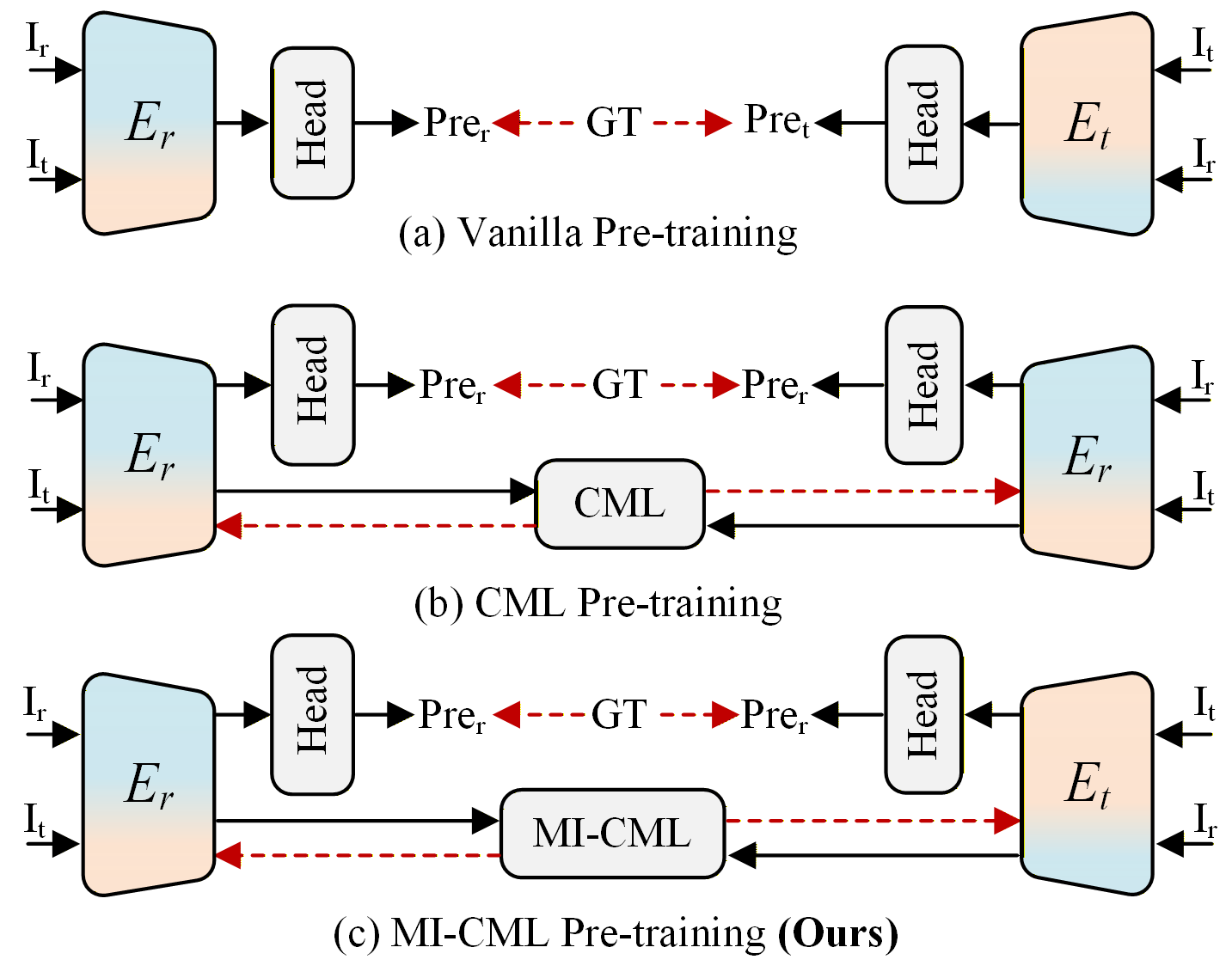}
      \caption{Three RGB-T pre-training strategies: (a) vanilla pre-training, (b) CML pre-training that utilizing the contrastive mutual learning to exchange knowledge between two identical RGB-dominant encoders (c) MI-CML pre-training \textbf{(ours)}.}
      \label{fig8}
\end{figure}

\begin{table}[t]
\caption{{Comparison of different RGB–T parameter ratios in the TUNI encoder.}}
\centering
\begin{tabular}{c|cc|cc|c}
\toprule
\rowcolor[gray]{.9}
RGB:T &Vanilla&MI-CML&Params&FLOPs&mIoU\\
\midrule
1:1   &\ding{52}     &    -   &    16.78&  18.77  &78.2\\
1:1    &    - &\ding{52}      &        16.78&  18.77  &78.3\\
\midrule
2:1     &\ding{52}     &-&10.19       &  11.55  &77.3\\
\rowcolor{cyan!10}
2:1     &    - &\ding{52}&   10.19        &  11.55  &\textbf{78.6}\\
\midrule
4:1 &\ding{52}     &-&   8.11&   9.34        &76.2\\
4:1    &    - &\ding{52}&  8.11&9.34    &77.9\\
\bottomrule
\end{tabular}
\label{table9}
\end{table}

{In our setting, the RGB-T pre-training is based on pseudo-thermal data due to the lack of large-scale real thermal datasets. Although a discrepancy may exist between synthetic and real thermal modalities, pre-training still enables the learning of general cross-modal semantic representations. Moreover, the subsequent fine-tuning on real RGB-T datasets enables the model to adapt to real thermal distributions, thereby mitigating the synthetic-to-real discrepancy.}

\textit{\textbf{5) Effectiveness of MI-CML Pre-training.}} Fig. \ref{fig8} illustrates three pre-training strategies: (a) vanilla pre-training; (b) CML pre-training, where two identical RGB-biased encoders $E_r$ learn from each other; and (c) MI-CML pre-training, which enables mutual learning between the $E_r$ and  $E_t$ through modality inversion. {As shown in Table \ref{table8}, $E_r$ consistently outperforms $E_t$, likely due to the richer information available in the RGB modality.} Compared with vanilla pre-training results, MI-CML improves the mIoU by 1.4\% and 1.3\% on $E_t$ and $E_r$, respectively. This suggests that MI-CML alleviates modality bias during pre-training, thereby enhancing performance on RGB-T semantic segmentation. Moreover, CML pre-training fails to yield performance gains for $E_r$ in the segmentation task, indicating that modality inversion is a key factor in our pre-training strategy.

{To further investigate the relationship between modality bias and MI-CML, we conduct experiments under different RGB-T parameter settings in the TUNI encoder, including 1:1, 1:2, and 1:4. As shown in Table \ref{table9}, when the parameter ratio is set to 1:1, the model is least affected by modality bias, and MI-CML brings only marginal improvement. As the parameter allocation becomes increasingly imbalanced (e.g., 2:1 and 4:1), the impact of modality bias becomes more pronounced. In these cases, MI-CML consistently yields more significant performance gains, improving mIoU from 77.3\% to 78.6\% under the 2:1 setting and from 76.2\% to 77.9\% under the 4:1 setting. These results indicate that the effectiveness of MI-CML is closely related to the degree of modality bias. Considering the trade-off between performance and model efficiency, we adopt the 2:1 parameter ratio as our default setting.}

{The pre-training time and GPU memory consumption are summarized in Table \ref{table10}. Due to the use of the alternating mutual learning strategy, the GPU memory usage shows negligible increase, while the training time approximately doubles.}

\begin{table}[t]
\caption{{Pre-training time and GPU memory consumption of various pre-training strategies. The training is conducted on 8 NVIDIA H20 GPUs with a per-GPU batch size of 128.}}
\centering
\begin{tabular}{c|cc|cc}
\toprule
\rowcolor[gray]{.9}
Encoder                 &Vanilla        &MI-CML         & \makecell{Memory \\ (GB/GPU)}  & \makecell{Time \\ (h)}\\
\midrule
\multirow{2}{*}{TUNI-T}     &\ding{52}      &    -      &  23.1& 30\\
 &   -    &    \ding{52}    &  23.5& 48 \\
 \midrule
\multirow{2}{*}{TUNI-S}     &   \ding{52}   &     -      &  26.7  &36\\
     &   -   &    \ding{52}       &    28.1&62\\
\midrule
\multirow{2}{*}{TUNI-B}     &  \ding{52}   &  -         &  40.9&66 \\
     &   -  &    \ding{52}        &    42.9&120\\
\bottomrule
\end{tabular}
\label{table10}
\end{table}

\begin{table}[t]
\caption{Comparison of various fine-tuning strategies.}
\centering
\begin{tabular}{cccc|c}
\toprule
\rowcolor[gray]{.9}
MLP     &  Cat\&MLP      &2×MLP      &MRL   &mIoU\\
\midrule
\ding{52}        &   -    &      - &  -  &78.6\\
-     &\ding{52}      &           -&   - &78.6\\
-     &   -   &\ding{52}           &   - &78.2\\
\rowcolor{cyan!10}
-       & -      &-       &\ding{52}    &\textbf{79.7}\\
\bottomrule
\end{tabular}
\label{table11}
\end{table}

\begin{figure}[t!]
      \centering
      \includegraphics[width=0.8\columnwidth]{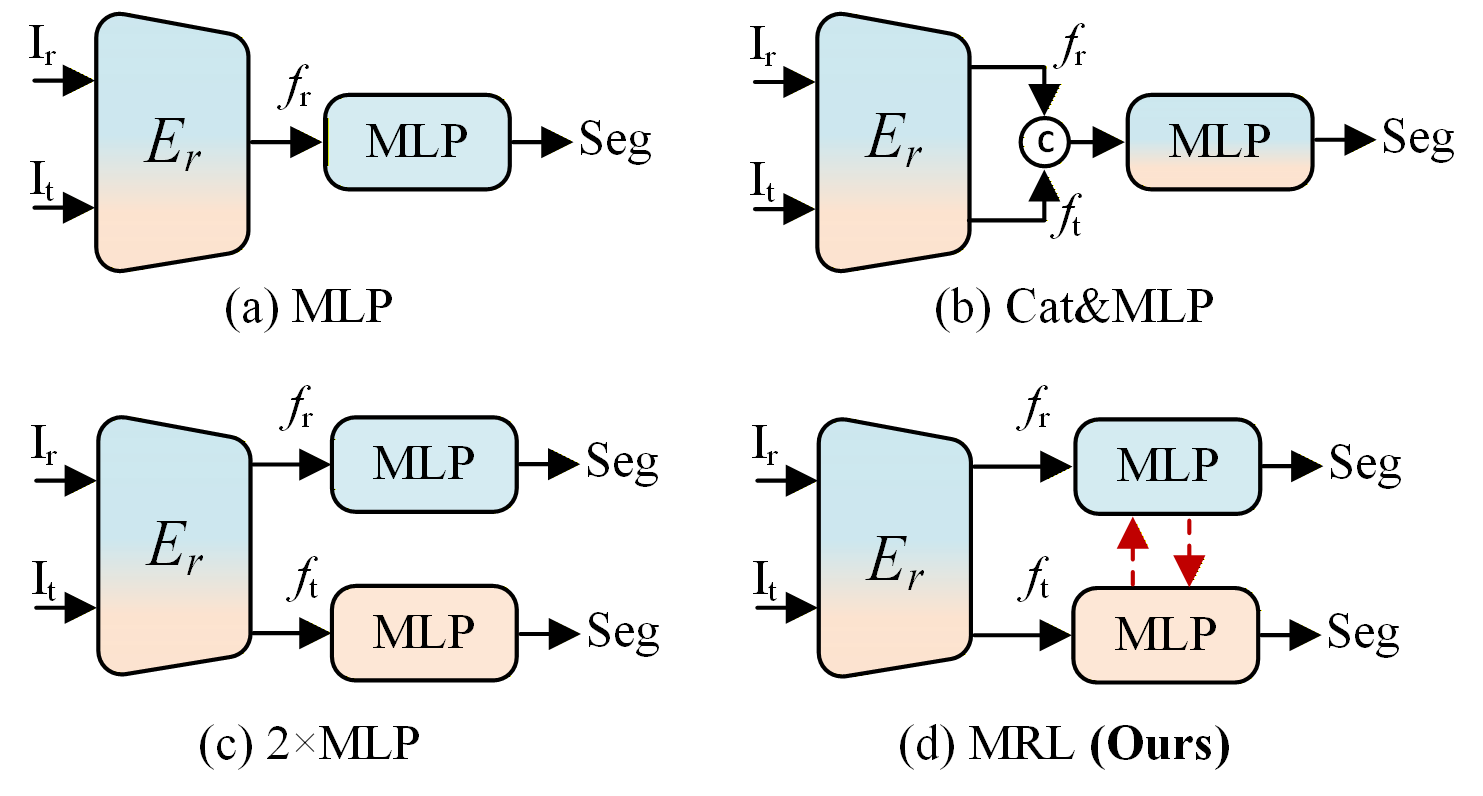}
      \caption{Four segmentation fine-tuning strategies: (a) using a single MLP to decode RGB features, (b) concatenating RGB and thermal features before a single MLP decoder, (c) employing two separate MLP decoders for RGB and thermal features with joint optimization, and (d) MRL fine-tuning \textbf{(Ours)}.}
      \label{fig9}
\end{figure}

\textit{\textbf{6) Effectiveness of MRL fine-tuning.}} Table \ref{table11} provides a preliminary validation of the effectiveness of MRL. Here, we perform a detailed analysis of its impact through comparison with three variants, thereby validating the motivation behind our study. As shown in Fig. \ref{fig9}, ‘MLP’ uses a single MLP to decode RGB features, ‘Cat\&MLP’ concatenates RGB and thermal features before a single MLP decoder, and ‘2×MLP’ employs two separate MLP decoders for RGB and thermal features with joint optimization. Compared with ‘MLP’, ‘Cat\&MLP’ yields no performance gain in segmentation, whereas ‘2×MLP’ leads to a slight decrease. {The former may stem from TUNI encoder already fusing RGB and thermal features, such that direct concatenation can overwhelm the thermal residual information; the latter may result from the difficulty of jointly optimizing two decoders. In contrast, MRL improves the mIoU by 1.1\%, demonstrating its effective use of thermal residual information by transferring complementary cues to the RGB branch during fine-tuning, enabling strong performance with only the RGB decoder at inference.}

\begin{table}[t]
\centering
\caption{{Effect of the temperature parameter $\tau$ under different $\lambda=\beta$ settings during MI-CML pre-training. The parameter $\gamma$ is fixed to 1, which corresponds to removing the MRL. Results are reported in mIoU.}}
\begin{tabular}{c|ccccc}
\toprule
\rowcolor[gray]{.9}
\diagbox{$\lambda=\beta$}{$\tau$} 
& 0.01 & 0.07 & 0.5 & 1 & 2\\
\midrule
0.5 & 77.6 & 77.7 & 77.4 & 77.4 & 76.9 \\
0.7 & 78.2 & \underline{78.4} & 77.8  & 77.4 & 77.0 \\
0.9 & 78.2 & \textbf{78.6} & 78.1 & 77.8 & 77.6 \\
\bottomrule
\end{tabular}
\label{table12}
\end{table}



\begin{figure}[t!]
      \centering
      \includegraphics[width=1.0\columnwidth]{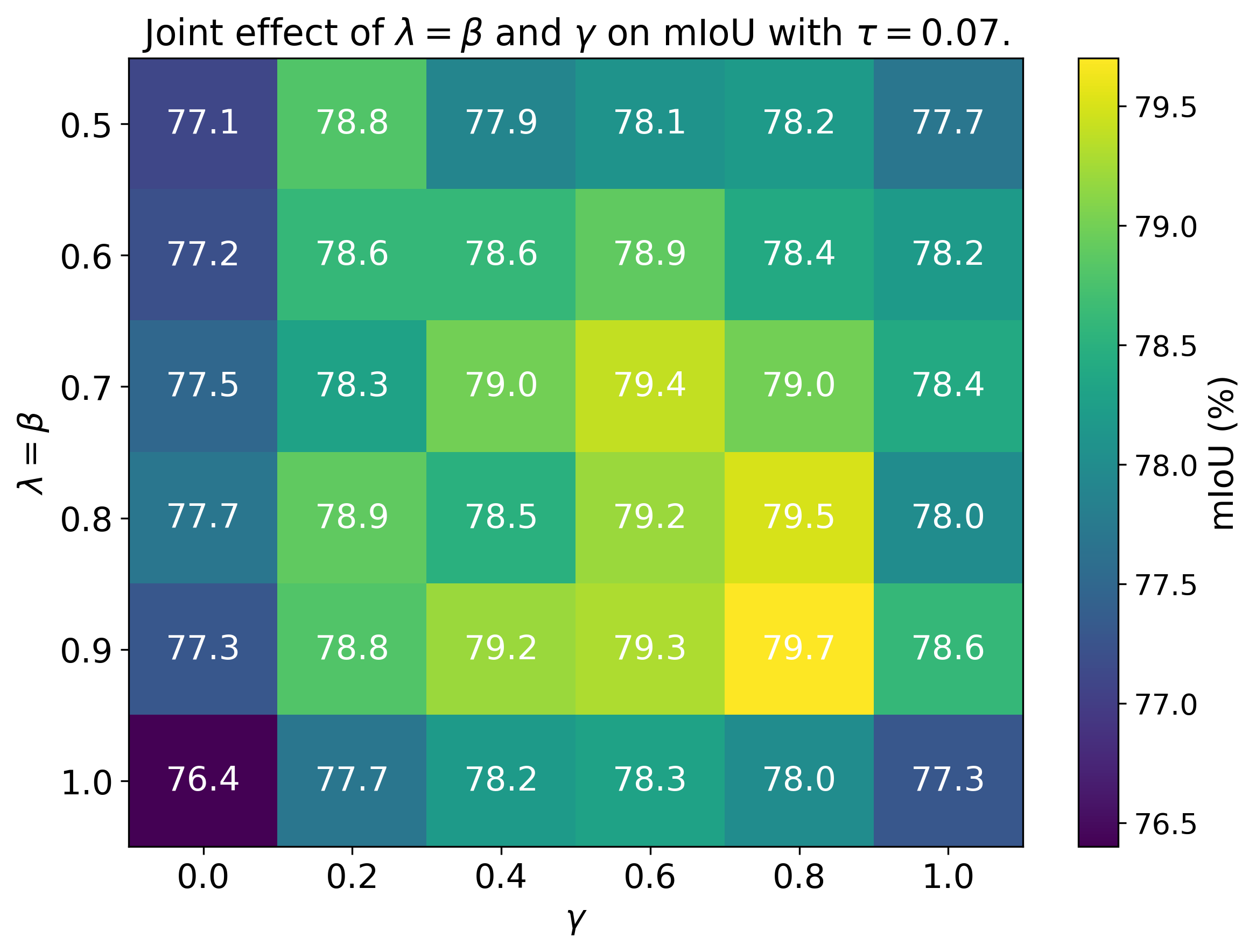}
      \caption{{Heatmap of the joint effects of $\lambda=\beta$ and $\gamma$ on mIoU with $\tau=0.07$, where brighter colors indicate higher values.}}
      \label{fig10}
\end{figure}

{\textit{\textbf{7) Sensitivity analysis of hyperparameters.}} {We conduct hyperparameter sensitivity experiments to investigate the effects of $\tau$, $\lambda$, $\beta$, and $\gamma$ on the TUNI framework. Since $\lambda$ and $\beta$ are conceptually symmetric (Eqs. (14)--(15)), they are set to the same value in the analysis. We first study the optimal temperature $\tau$ during pre-training under three $\lambda=\beta$ settings, with $\gamma$ fixed to 1 to remove MRL and isolate the effect of MI-CML. As shown in Table XII, $\tau = 0.07$ consistently achieves the best performance, while $\tau = 0.01$ yields comparable results. This is consistent with common observations in contrastive learning, where a relatively small temperature produces sharper contrastive distributions. As $\tau$ increases, the effectiveness of MI-CML gradually diminishes. Therefore, we set the default temperature to 0.07.}

{Fig. 10 further illustrates the joint effects of $\lambda=\beta$ and $\gamma$ with $\tau=0.07$. TUNI performs better when $\lambda=\beta$ is in the range of 0.7--0.9 and $\gamma$ is around 0.6--0.8, indicating that MI-CML and MRL should be properly balanced. These results suggest that supervised learning and the RGB branch tend to dominate pre-training and fine-tuning, respectively, while MI-CML alleviates modality bias and the thermal branch provides supplementary cues. Based on these observations, we set $\lambda=\beta=0.9$ and $\gamma=0.8$ as the default hyperparameter settings, which achieve the best performance.}

\subsection{Limitation and Future Work}
Although our work addresses several challenges in prevailing RGB-T semantic segmentation frameworks, {there remain several limitations.} First, the TUNI encoder exhibits diminishing marginal performance gains as the network scale increases. In our experiments, doubling the scale of TUNI-B yields less than a 1\% improvement in segmentation performance. These results suggest that simply scaling up the TUNI encoder is insufficient and that architectural optimization is necessary to achieve further performance improvements. Second, the pseudo-thermal images generated via image translation differ from real thermal images, which may constrain further improvement of the framework. Advances in RGB-T image translation techniques could potentially enhance the framework’s performance. Third, our pre-training is conducted on paired RGB-T images, which poses challenges for generalization to unpaired application scenarios. Pre-training on unpaired RGB-T data could broaden the applicability of the framework.

\section{Conclusions}
{This paper presents TUNI, a unified pre-training and fine-tuning framework for real-time and efficient RGB-T semantic segmentation. The core idea of TUNI is to enhance the utilization of thermal information across both the pre-training and fine-tuning stages. Specifically, the TUNI encoder incorporates an R–T LM module to adaptively emphasize consistent and complementary local features across modalities, while the MI-CML pre-training strategy alleviates RGB bias through contrastive and mutual learning under modality inversion, promoting better thermal-aware representation learning. During fine-tuning, the MRL decoder further exploits correct yet divergent prediction regions between modality-specific decoders to effectively extract residual thermal information.} Extensive experiments on five RGB-T benchmarks demonstrate that TUNI achieves strong performance across different efficiency–accuracy trade-offs, where TUNI-B attains state-of-the-art results, TUNI-S delivers competitive accuracy with only 10M parameters and 120 FPS on an RTX 4090, and TUNI-T enables deployment on resource-constrained platforms with 5M parameters and 35 FPS on Jetson Orin NX. {Overall, TUNI provides a unified and effective solution for robust and efficient RGB-T semantic segmentation across diverse real-world scenarios.} Future work will explore enhanced encoder designs and pre-training on unpaired RGB-T data to further improve generalization.

\bibliographystyle{IEEEtran}
\bibliography{IEEEabrv,IEEEexample}

\IEEEbiographynophoto{Xiaodong Guo}
is currently pursuing a Ph.D. degree at the School of Automation, Beijing Institute of Technology. His research interests include cross-modal feature fusion, semantic segmentation, knowledge distillation, and object detection. 

\vspace{-1cm}
\IEEEbiographynophoto{Xianda Guo}
is currently a Ph.D. student at the School of Computer Science, Wuhan University. Before that, he received a master's degree from the Dalian University of Technology in 2019. His work is mainly in computer vision, such as depth estimation and stereo-matching. He has co-authored publications on TPAMI, CVPR, ICCV, and ECCV. He is the creator of the widely used open-source platform OpenStereo.

\vspace{-1cm}
\IEEEbiographynophoto{Tong Liu}
received the Ph.D. degree in navigation,
guidance and control from the Beijing Institute of
Technology, Beijing, China, in 2013.
He is currently an Associate Professor with the
Beijing Institute of Technology. His research interests mainly include inertial sensors and intelligent
navigation.

\vspace{-1cm}
\IEEEbiographynophoto{Zhihong Deng}
was born in Liaoning, China, in 1974. She received the B.S. and Ph.D. degrees in control theory and engineering from the Department of Automation, Harbin Engineering University, Harbin, China, in 1997 and 2001, respectively. From December 2001 to December 2003, she was a Postdoctoral Research Associate with the Beijing Institute of Technology, Beijing, China, and worked on navigation, guidance, and control. Since 2004, she has been an Associate Professor with the Department of Automatic Control, Beijing Institute of Technology, and then, as a Professor since 2010. Her current research interests include strap down inertial navigation systems, information fusion and filter, and intelligent navigation technology.
\vspace{-1cm}
\IEEEbiographynophoto{Yanlun Peng}
is currently the Director of Autonomous Driving R\&D at Great Wall Motor. Prior to this role, he accumulated 2 years of professional experience at Horizon Robotics. With a proven track record in end-to-end engineering implementation, he has spearheaded the development and deployment of a comprehensive autonomous driving technology stack, spanning from Occupancy Network (Occ) one-stage end-to-end solutions to cutting-edge Vehicle-Level Autonomy (VLA) systems. His work focuses on bridging advanced algorithmic research with practical industrial applications, driving the mass production and iteration of autonomous driving technologies.
\vspace{-1cm}
\IEEEbiographynophoto{Xiang Li}
is currently pursuing the M.S. degree at Beijing Institute of Technology, Zhuhai, China. His research interests include image fusion, image enhancement, and multimodal semantic segmentation.
\vspace{-1cm}
\IEEEbiographynophoto{Wujie Zhou} is currently a Professor with the School of Information and Electronic Engineering, Zhejiang University of Science and Technology, Zhejiang, China. Previously, he served as a Post-Doctoral Fellow at the Institute of Information and Communication Engineering, Zhejiang University in Zhejiang. He is a Visiting Scholar at Nanyang Technological University in Singapore. He has authored more than 70 technical articles in refereed journals and conference proceedings in the fields of artificial intelligence and multimedia signal processing, including publications in prestigious venues such as AAAI, IEEE TIP, IEEE TMM, IEEE TCSVT, IEEE TNNLS, IEEE TITS, IEEE JSTSP, IEEE TSMC, IEEE TGRS, IEEE TBC, IEEE TIV, IEEE JSTARS, IEEE TCI, IEEE TIM, IEEE MIS, IEEE TCDS, IEEE TETCI, PR, and Information Fusion. These articles include more than 30 published in IEEE Journals/Transactions/Magazines, 7 designated as ESI hot papers, 14 recognized as ESI highly cited papers, 4 with more than 100 citations, and more than 10 selected as TOP 50 Popular Articles in IEEE TIP, IEEE TCSVT, IEEE TMM, IEEE MIS, and IEEE TETCI. In October 2022, he was ranked among the top 2\% of the world's leading scientists on the list released by Stanford University. His research interests primarily focus on image processing and computer vision. Additionally, he serves as an Associate Editor for Mathematics and Frontier in Computational Neuroscience. He has been invited to review papers for prestigious conferences such as AAAI and CVPR, as well as several IEEE journals including TIP, TMM, TCSVT, TNNLS, JSTSP, TSMC, TBC, and TIM.

\end{document}